\documentclass{article}

\usepackage{microtype}
\usepackage{graphicx}
\usepackage{subfigure}
\usepackage{booktabs} 

\usepackage{epsfig}
\usepackage{amssymb}
\usepackage{amsfonts}
\usepackage{tikz}
\usepackage{float}
\usepackage{bbm}
\usepackage{dsfont}
\usepackage{lineno}
\usepackage{amsmath}
\usepackage{amsthm}
\usepackage{balance}
\usepackage{lineno}
\usepackage{physics}
\usepackage{courier}
\usepackage[hyphens]{url}

\usepackage{hyperref}

\usepackage[accepted]{icml2018x}

\icmltitlerunning{Distributionally Robust Removal of Malicious Nodes from Networks}

\newcommand*\circled[1]{\tikz[baseline=(char.base)]{
            \node[shape=circle,draw,inner sep=0.5pt] (char) {#1};}}

\newtheorem{theorem}{Theorem}
\newenvironment{customtheorem}[1]
{\renewcommand\thetheorem{#1}\theorem}
{\endtheorem}

\newtheorem{lemma}{Lemma}[section]
\newenvironment{customlemma}[1]
  {\renewcommand\thelemma{#1}\lemma}
  {\endlemma}

\allowdisplaybreaks

\begin{document}
\twocolumn[
\icmltitle{Distributionally Robust Removal of Malicious Nodes from Networks}




\begin{icmlauthorlist}
\icmlauthor{Sixie Yu}{wustl}
\icmlauthor{Yevgeniy Vorobeychik}{wustl}
\end{icmlauthorlist}

\icmlaffiliation{wustl}{Computer Science and Engineering Department, Washington University in St. Louis}

\icmlcorrespondingauthor{yvorobeychik@wustl.edu}

\icmlkeywords{Machine Learning, ICML}

\vskip 0.3in
]



\printAffiliationsAndNotice{}  

\begin{abstract}
An important problem in networked systems is detection and removal of suspected
malicious nodes.
A crucial consideration in such settings is the uncertainty
endemic in detection, coupled with considerations of network
connectivity, which impose indirect costs from mistakely removing
benign nodes as well as failing to remove malicious nodes.
A recent approach proposed to address this problem directly tackles
these considerations, but has a significant limitation: it assumes
that the decision maker has accurate knowledge of the joint
maliciousness probability of the nodes on the network.
This is clearly not the case in practice, where such a distribution is at
best an estimate from limited evidence.
To address this problem, we propose a distributionally robust 
framework for optimal node removal.
While the problem is NP-Hard, we propose a principled algorithmic
technique for solving it approximately based on duality combined with Semidefinite
Programming relaxation.
A combination of both theoretical and empirical analysis, the latter using both
synthetic and real data, provide strong evidence that our algorithmic
approach is highly effective and, in particular, is significantly more robust than the state of the art.
\end{abstract}
\vspace{-0.1in}

\section{Introduction}\label{S:intro}
One of the major problems in networked settings is to identify and remove potentially malicious nodes.
For example, in social networks, malicious nodes may correspond to accounts created by malicious parties which spread social spam, hate speech, fake news, and the like, with considerable deliterious effects~\cite{allcott2017social,cheng2015antisocial}.
Major social network platforms consequently devote considerable efforts to identify and remove fake or malicious accounts~\cite{facebook18,nyt2017facebook}.
Nevertheless, evidence suggests that the problem remains pervasive~\cite{vinicius2018brazil,narayanan2018polarization}.
Similarly, in cyber-physical systems (e.g., smart grid infrastructure), computing nodes compromised by malware can cause catastrophic losses, and mitigation through detection and removal of such malicious nodes is a major problem~\cite{mo2012cyber, Yang17}.

A central challenge faced in deciding which potentially malicious nodes to remove is to account for the combination of uncertainty about whether particular nodes are malicious, and the indirect (network) effects of the decision.
This combination makes the decision about which nodes to remove fundamentally a  subset selection problem---a challenging combinatorial optimization problem.
Recently, \citeauthor{yu2018removing} proposed an approach for solving it they term \texttt{MINT}, where the problem is captured by approximately minimizing loss which involves three terms: direct loss from removing benign nodes, indirect loss from cutting links in the benign subgraph, and indirect loss from maintaining connectivity between malicious and benign nodes.
\begin{figure}[h]
\centering
\setlength{\tabcolsep}{0.1pt}
\begin{tabular}{c}
\includegraphics[width=2.5in]{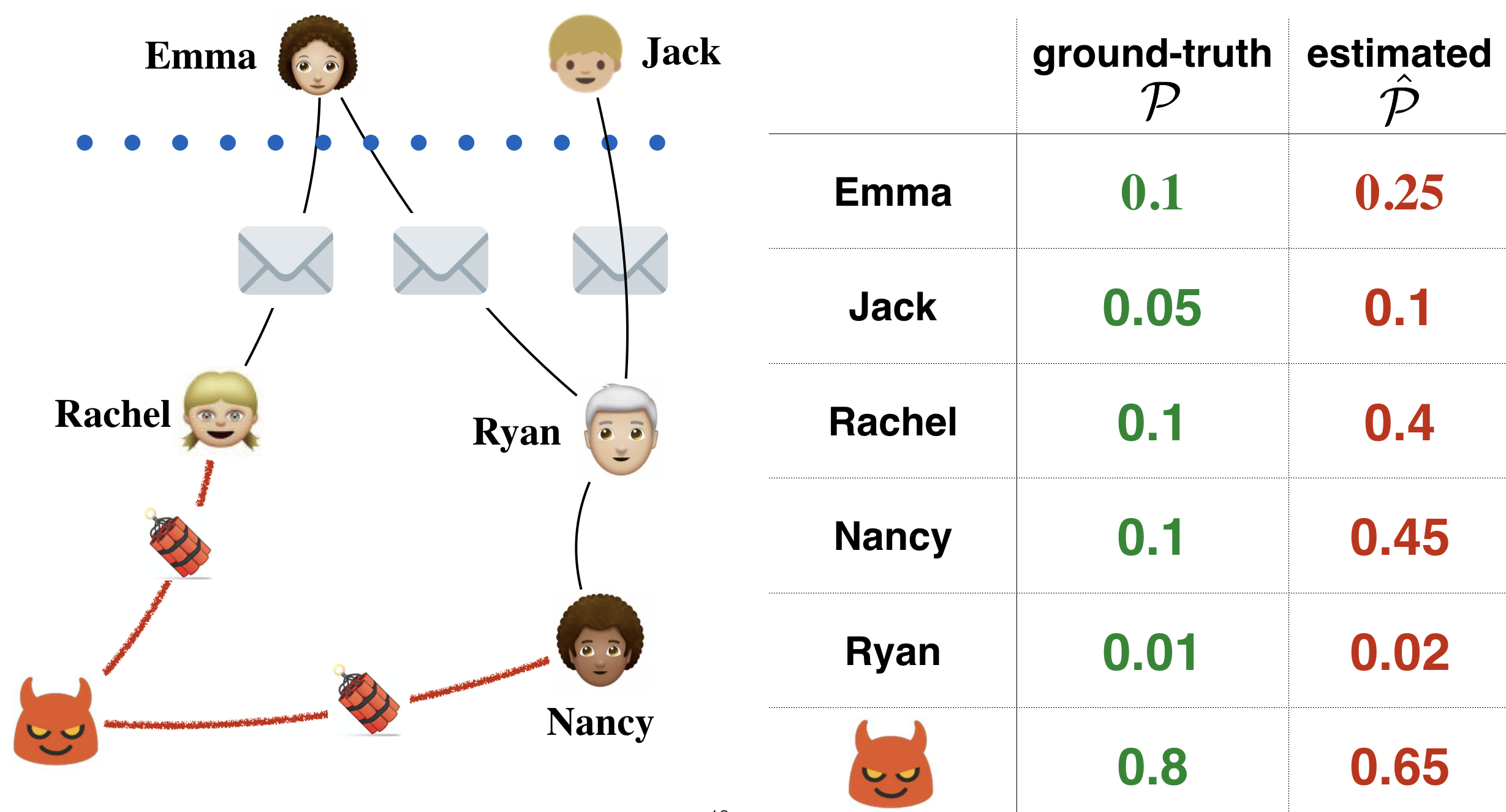}
\end{tabular}
\caption{An illustration of a decision to remove two nodes, Jack and Emma, from the network, on our loss function.}
\label{fig:demo}
\end{figure}
This model is illustrated in Fig.~\ref{fig:demo}, where we 
consider removing Jack and Emma, two \textit{benign} nodes above the dotted blue line (and failing to remove the malicious node).
Suppose that we pay a penalty of $\alpha_1$ for each benign node we remove, a penalty $\alpha_2$ for each link we cut between benign nodes, and $\alpha_3$ for each link between \textit{remaining} malicious nodes and benign nodes. 
Since we remove $2$ benign nodes, cut $3$ links between benign nodes (one between Emma and Rachel, one between Emma and Ryan, and another between Jack and Ryan), and the malicious node is still connected to $2$ nodes (Rachel and Nancy), our total loss is: $2\alpha_1 + 3\alpha_2 + 2\alpha_3$. 

A major shortcoming of \texttt{MINT} is that it assumes that the distribution of node maliciousness is known.
In practice, such a distribution is estimated from limited evidence, such as node behavior and other characteristics, and this estimation may be quite inaccurate (particularly if our modeling assumptions are poor, for example, if we erroneously assume that maliciousness probabilities of nodes are independent).
More precisely, consider an unknown \textit{ground-truth} $\mathcal{P}$, as illustrated in Fig.~\ref{fig:demo} in green.
Whereas \texttt{MINT} assumes we know $\mathcal{P}$, in reality we only have an estimate $\hat{\mathcal{P}}$ (shown in red in Fig.~\ref{fig:demo}).
To address this issue, we propose a new approach, \texttt{MINT\_DRO}, which is a  distributionally robust framework for optimal node removal. 
We design an uncertainty set around the estimate $\hat{\mathcal{P}}$ and optimize with respect to the \textit{worst-case} scenario. 
We propose a principled algorithmic approach for solving this problem approximately based on duality combined with Semidefinite Programming relaxation, and prove that the uncertainty set in our model contains the ground-truth distribution $\mathcal{P}$ with high probability.
This in turn implies that with high probability \texttt{MINT\_DRO} is robust with respect to the ground-truth distribution. 
Finally, we conducted extensive experiments using both synthetic and real data to show that our model is significantly more robust than \texttt{MINT}.  


\paragraph{Related Work} 

There are several prior efforts considering a related problem of \textit{graph scan statistics} and hypothesis testing~\cite{arias2011detection,priebe2005scan,sharpnack2013near}. These study the following problem: given a graph $G$ where each node is associated with a random variable with an exogenously specified probability distribution, find a subset of nodes that maximizes a scan statistic defined over subsets of nodes (for example, this statistic may generalize log-likelihood ratio).
The recent \texttt{MINT} approach~\cite{yu2018removing} can be viewed through this lens as well, but as it has been shown to have state-of-the-art performance, our comparison, our experimental evaluation focuses on comparing to \texttt{MINT}.

Also closely related to our problem is the broader literature on distributionally robust optimization (DRO) \cite{scarf1958min}. 
In the DRO framework one defines a set of probability distributions that is assumed to contain the true stochastic model of the problem. Many solutions have been proposed to solve specific problems under the DRO framework~\cite{xu2010distributionally,calafiore2006distributionally,yue2006expected,cheng2014distributionally,wiesemann2014distributionally}, although this framework has not been applied in the context of choosing which potentially malicious nodes to remove from a network.

Our design of the uncertainty set is inspired by the idea of moment-constrained uncertainty set~\cite{delage2010distributionally,popescu2007robust,calafiore2006distributionally}. 
Yet another related research strand is in using Semidefinite Programming (SDP) to approximate combinatorial optimization problems~\cite{goemans1995improved,luo2010semidefinite,bertsimas2000moment}, although such approaches are domain specific.
Finally, our work bears some relationship to the burgeoning field of adversarial machine learning~\cite{Vorobeychik18}, although we do not explicitly consider issues of adversarial response (such as evasion attacks) in our setting.

\section{Model}\label{S:model}
We consider a network that is represented by a graph $G=(V, E)$, where $V$ ($|V|=N$) is the set of nodes and $E$ the set of edges connecting them. Each node $i \in V$ represents a user and each edge $(i, j)$ represents an edge (e.g., friendship on Facebook) between user $i$ and user $j$. We focus our attention on undirected graphs. We denote the adjacency matrix of $G$ by $\mathbf{A}\in \mathbb{R}^{N \times N}$. The elements of $\mathbf{A}$ are binary if the graph is unweighted, or some non-negative real numbers if the graph is weighted. To make expositioin easier we focus on unweighted graphs. Generalization to weighted graphs is straightforward.

We consider the problem of removing malicious nodes from the network $G$.  A \textit{configuration} of the network is denoted by $\mathbf{\pi } \in \{ 0, 1 \}^N$, with $\pi _i=1$ indicating that a node $i$ is malicious, with $\pi _i=0$ when $i$ is benign. For convenience, we also let $\bar{\pi }_i=1 - \mathbf{\pi }_i$ to indicate that $i$ is benign. Consequently, $\mathbf{\pi }$ (and $\mathbf{\bar{\pi }}$) assigns malicious or benign label to each node. The identity of malicious and benign nodes are usually uncertain. So instead we have a probability distribution over the configurations. Formally, let $\mathbf{\pi } \sim \mathcal{P}$, where $\mathcal{P}$ captures the joint probability distribution over node configurations.

Our work builds upon the following model proposed by \citet{yu2018removing}.
Let $S$ denote the set of nodes to remove. Define a vector $\mathbf{x} \in \{ -1, 1 \}^N$, where $\mathbf{x}_i=1$ if and only if node $i$ is removed ($i \in S$), and $\mathbf{x}_i=-1$ if node $i$ remains in the network ($i \in V \setminus S$). The goal of their model is to identify a subset of nodes $S$ to remove so as to minimize the impact of the remaining malicious nodes on the network, while at the same time minimizing disruptions caused to the benign subnetwork. This goal is naturally captured by the loss function given in~Eq.~\eqref{eq:expected_loss}.
\vspace{-0.1in}
\begin{align}\label{eq:expected_loss}
\begin{split}
& \mathcal{L}(\mathbf{x}) := \\
& \alpha_1 \underbrace{\sum_{i=1}^{N}{\mathbf{x}_i \mathbb{E}_{\pi  \sim \mathcal{P}}[  \bar{\mathbf{\pi }}_i}]}_{\mathcal{L}_1}   - \alpha_2  \underbrace{\sum_{i,j}^{N}{\mathbf{A}_{i,j} \mathbf{x}_i \mathbf{x}_j \mathbb{E}_{\pi  \sim \mathcal{P}}[ \bar{\mathbf{\pi }}_i \bar{\mathbf{\pi }}_j}]}_{\mathcal{L}_2} + \\ 
&  \quad \, \alpha_3  \underbrace{\sum_{i,j}^{N}{\mathbf{x}_i \mathbf{x}_j \mathbf{A}_{i,j} \mathbb{E}_{\pi  \sim \mathcal{P}}[ \mathbf{\pi }_i \mathbf{\bar{\pi }}_j}] }_{\mathcal{L}_3}.
\end{split}
\end{align}
As we can observe, the loss function is composed of three components.
The first component, $\mathcal{L}_1$, of the loss function is the direct loss associated with removing benign nodes.
The second component, $\mathcal{L}_2$, penalizes cutting connections between benign nodes that are removed and benign nodes that remain; in other words, it penalizes the degradation of connectivity within the benign subgraph. The third component of the loss function, $\mathcal{L}_3$, captures the consequence of failing to remove malicious nodes in terms of connections from these to benign nodes. The  nonnegative trade-off parameters $\alpha_1$, $\alpha_2$, and $\alpha_3$ satisfy $\alpha_1 + \alpha_2 + \alpha_3=1$, and weigh the relative importance of the three components of the loss function. 

The configuration $\mathbf{\pi }$ is a random variable distributed according to $\mathcal{P}$. Let $\mathbf{\mu}\in \mathbb{R}^N$ and $\mathbf{\Sigma}\in \mathbb{R}^{N \times N}$ denote its mean and covariance, respectively. The loss function defined in Eq.~\eqref{eq:expected_loss} depends on both $\mathbf{\mu}$ and $\mathbf{\Sigma}$. To make the dependency explicit we define several matrices and re-write the loss function in a matrix-vector form. We define the matrices $\mathbf{B}(\mathbf{\mu}), \mathbf{P}(\mathbf{\mu}, \mathbf{\Sigma}), \mathbf{M}(\mathbf{\mu}, \mathbf{\Sigma})$ as follow.\footnote{$diag(\mathbf{x})$ returns a diagonal matrix with diagonal elements equal to $\mathbf{x}$.} 
\vspace{-0.1in}
\begin{equation*}
\begin{aligned}
\mathbf{B}(\mathbf{\mu})  :&= diag\big( \mathbb{E}_{\mathbf{\pi} \sim \mathcal{P}}[\bar{\mathbf{\pi}}] \big) \\ 
\mathbf{P}(\mathbf{\mu}, \mathbf{\Sigma})  :&= \mathbf{A} \odot \mathbb{E}_{\mathbf{\pi} \sim \mathcal{P}}[\bar{\mathbf{\pi}} \bar{\mathbf{\pi}}^T] \\
\mathbf{M}(\mathbf{\mu}, \mathbf{\Sigma})  :&= \mathbf{A} \odot \mathbb{E}_{\mathbf{\pi} \sim \mathcal{P}}[ \mathbf{\pi} \bar{\mathbf{\pi}}^T] 
\end{aligned}
\end{equation*}
Note that the elements of these matrices are not constant, but depend on $\mathbf{\mu}$ and $\mathbf{\Sigma}$ (see the appendix for their detailed dependency).

Slightly abusing notation, we define two additional matrices, $\mathbf{Q}(\mathbf{\mu}, \mathbf{\Sigma})$ and $\mathbf{b}(\mathbf{\mu})$. Note that $\mathbf{Q} \in \mathbb{R}^{N \times N}$ is a symmetric matrix: 
\vspace{-0.1in}
\begin{equation*}
\begin{aligned}
\mathbf{Q}(\mathbf{\mu}, \mathbf{\Sigma}):= & (\alpha_3/2)\big[\mathbf{M}(\mathbf{\mu}, \mathbf{\Sigma}) + \mathbf{M}(\mathbf{\mu}, \mathbf{\Sigma})^T   \big] - \\
								           & (\alpha_2/2)\big[ \mathbf{P}(\mathbf{\mu}, \mathbf{\Sigma}) + \mathbf{P}(\mathbf{\mu}, \mathbf{\Sigma})^T  \big],
\end{aligned}
\end{equation*}
and $\mathbf{b}(\mathbf{\mu}):=(\alpha_1/2) \mathbf{B}(\mathbf{\mu}) \mathbf{1}$.
We can now rewrite the loss function in a compact matrix-vector form as the following: 
\vspace{-0.1in}
\begin{equation*}
\begin{aligned}
\mathcal{L}(\mathbf{x}; \mathbf{\mu}, \mathbf{\Sigma}) &  = \mathbb{E}_{\mathbf{\pi } \sim \mathcal{P}} \big[\alpha_1 \mathcal{L}_1 + \alpha_2 \mathcal{L}_2 + \alpha_3 \mathcal{L}_3 \big] \\
					    & = \mathbf{x}^T \mathbf{Q}(\mathbf{\mu}, \mathbf{\Sigma}) \mathbf{x} + 2 \mathbf{x}^T \mathbf{b}(\mathbf{\mu})
\end{aligned}
\end{equation*}

Optimizing the loss function above (as done by \citeauthor{yu2018removing}) critically assumes that the maliciousness distribution $\mathcal{P}$ is known.
In reality, this is typically not the case, and such a distribution is estimated from data.
Let $\hat{\mathcal{P}}$ denote the estimated distribution. 
The mean of $\hat{\mathcal{P}}$ is denoted by $\hat{\mathbf{\mu}}$, where $\hat{\mathbf{\mu}}_i$ is the estimated probability that node $i$ is malicious given its features from past data. Similarly, the estimated covariance matrix is represented by $\hat{\mathbf{\Sigma}}$.  The model proposed by~\citeauthor{yu2018removing} is called \texttt{MINT}, which is to solve the following optimization problem:
\vspace{-0.1in}
 \begin{equation}\label{eq:MINT_opt}\tag{\texttt{MINT}}
\begin{aligned}
& \min_{\mathbf{x}} & & \mathcal{L}(\mathbf{x}; \hat{\mathbf{\mu}}, \hat{\mathbf{\Sigma}})  \\
&s.t.     &    &    \mathbf{x} \in \{ -1,1\}^N
\end{aligned}
\end{equation}
\vspace{-0.1in}

Although \texttt{MINT} has been shown to perform well on several real-world datasets, its performance is strongly influenced by the estimation error of $\mathbf{\mu}$. 
In fact, in Section~\ref{S:exp} we show that even a small estimation error can severely undermine the performance of \texttt{MINT}. 

In order to  mitigate the sensitivity of \texttt{MINT} to estimation error, we propose a novel \textit{Distributionally Robust Optimization} (DRO) approach for solving the problem posed above. 
The general idea is to design a distributional set to capture the uncertainty about the estimated mean $\hat{\mathbf{\mu}}$ and make decisions considering the \textit{worst-case} scenario. Specifically, we propose a model named \texttt{MINT\_DRO}, which aims to solve the following optimization problem:
\vspace{-0.1in}
 \begin{equation}\label{eq:MINT_DRO}\tag{\texttt{MINT\_DRO}}
\begin{aligned}
& \min_{\mathbf{x}} \sup_{\mathcal{F} \sim \Pi } & &  \mathbb{E}_{\mathcal{F}}\big[ \mathcal{L}(\mathbf{x}; \mathbf{\mu}_\mathcal{F}, \hat{\mathbf{\Sigma}}) \big] \\
&s.t.     &    &    \mathbf{x} \in \{ -1, 1\}^N,
\end{aligned}
\end{equation}
where the set $\Pi$ captures uncertainty about the true mean $\mathbf{\mu}$. 
There are several fundamental differences between \texttt{MINT\_DRO} and \texttt{MINT}. First, there is an additional inner maximization problem in \texttt{MINT\_DRO}. The inner maximization is optimized over  a set $\Pi $, which contains a set of probability distributions, where $\mathcal{F}$ is any distribution sampled from $\Pi$, and $\mathbf{\mu}_\mathcal{F}$ are random variables distributed according to $\mathcal{F}$. 
Inspired by~\citet{delage2010distributionally} and \citet{cheng2014distributionally}, we parametrize the set $\Pi $ by the first and second moments of the distributions in it. Specifically, let $\mathcal{F}$ be any distribution in $\Pi $. 
Consider the following two constraints:
\vspace{-0.1in}
\begin{align}\label{eq:moment_const}
\begin{split}
& (\mathbb{E}[\mathbf{\mu}_\mathcal{F}] - \hat{\mathbf{\mu}})^T \hat{\mathbf{\Sigma}}^{-1} (\mathbb{E}[\mathbf{\mu}_\mathcal{F}] - \hat{\mathbf{\mu}}) \le \gamma_1 \\
& \mathbb{E}\big[(\mathbf{\mu}_\mathcal{F} - \hat{\mathbf{\mu}})(\mathbf{\mu}_\mathcal{F} - \hat{\mathbf{\mu}})^T \big] \preceq \gamma_2 \hat{\mathbf{\Sigma}},
\end{split}
\end{align}
where $\hat{\mathbf{\mu}}$ and $\hat{\mathbf{\Sigma}}$ are the mean and covariance matrix estimated from data. $\mathbf{\mu}_\mathcal{F}$ are random variables distributed according to $\mathcal{F}$.
The first constraint defines an ellipsoid, which indicates that 
the expectation of $\mathcal{F}$ lies in the ellipsoid centered at the estimate $\hat{\mathbf{\mu}}$. The size of this ellipsoid is determined by $\gamma_1$, which provides a natural measure to quantify our uncertainty about $\mathbf{\mu}$ given $\hat{\mathbf{\mu}}$. Note that the second constraint also defines the support of the distribution $\mathcal{F}$.  The second constraint enforces the covariance matrix of $\mathcal{F}$ to lie in a positive semi-definite cone. Intuitively, the second constraint captures how likely it is that the random variable $\mathbf{\mu}_\mathcal{F}$ is close to $\hat{\mathbb{\mu}}$. The set $\Pi $ is then characterized by Eq.~\eqref{eq:uncertainty_set}:
\begin{align}\label{eq:uncertainty_set}
\begin{split}
 \Pi (\hat{\mathbf{\mu}}, \hat{\mathbf{\Sigma}}, & \gamma_1, \gamma_2):=  \\
& \Bigg\{ \mathcal{F} \Bigg| 
\begin{array}{ll}
(\mathbb{E}[\mathbf{\mu}_\mathcal{F}] - \hat{\mathbf{\mu}})^T \hat{\mathbf{\Sigma}}^{-1} (\mathbb{E}[\mathbf{\mu}_\mathcal{F}] - \hat{\mathbf{\mu}}) \le \gamma_1 \\
\mathbb{E}\big[(\mathbf{\mu}_\mathcal{F} - \hat{\mathbf{\mu}})(\mathbf{\mu}_\mathcal{F} - \hat{\mathbf{\mu}})^T \big] \preceq \gamma_2 \hat{\mathbf{\Sigma}}
\end{array}
  \Bigg\}
\end{split}
\end{align}
  
The set $\Pi$ is always non-empty, since it must contain the  distribution $\hat{\mathcal{P}}$. In Section~\ref{S:th} we provide probabilistic arguments to show that  $\Pi$ contains ground-truth distribution $\mathcal{P}$ with high probability, which guarantees that with high probability our model \texttt{MINT\_DRO} is robust with respect to the ground-truth distribution $\mathcal{P}$. 
The choice of the two  parameters $\gamma_1$ and $\gamma_2$ is important for  the robustness of \texttt{MINT\_DRO}. 
If their values are too small the benefit from the distributionally robust formulation is limited. In the extreme case where $\gamma_1$ and $\gamma_2$ are zeros our model \texttt{MINT\_DRO} reverts to \texttt{MINT}. 
On the other hand if their values are too large, our model would make excessively conservative decisions. In Section~\ref{S:th} we show how to make sensible choice of these values. 

\section{Solution Approach}
\label{S:solution}
In this section we derive the algorithm to solve our model \texttt{MINT\_DRO}. The optimization problem of \texttt{MINT\_DRO} is a binary quadratic program, which is  diffcult to solve even if the loss function $\mathcal{L}(\mathbf{x}; \mathbf{\mu}, \mathbf{\Sigma})$ is convex. Additionally, in our problem the loss function is nonconvex since the matrix $\mathbf{Q}$ is usually not positive (semi)-definite, further complicating the situation. 
Indeed, given that MINT, which was shown by \citeauthor{yu2018removing} to be NP-Hard, is a special case, the following result is immediate.
\begin{theorem}
Solving \texttt{MINT\_DRO} is NP-Hard. 
\end{theorem}

In what follows, we derive an approximation approach for solving \texttt{MINT\_DRO}. We first  apply duality to transform the inner maximization into a minimization problem, which can be jointly minimized with the outer minimization over $\mathbf{x}$. At this stage the optimization problem is still a NP-hard combinatorial optimization problem. Next, we apply  Semidefinite Programming (SDP) to obtain a convex relaxation of our problem which can be solved efficiently.

The support of the distributions in $\Pi$ is  $\mathcal{S}$, which is defined as $\mathcal{S}:=\big\{ \mathbf{\mu}_{\mathcal{F}} \, \big| \,   (\mathbf{\mu}_\mathcal{F} - \hat{\mathbf{\mu}})^T \hat{\mathbf{\Sigma}}^{-1} (\mathbf{\mu}_\mathcal{F} - \hat{\mathbf{\mu}}) \le \gamma_1 \big\}$, 
where the subscript of $\mathbf{\mu}_\mathcal{F}$ indexes the distribution associated with this random variable. Note that $\mathbf{\mu}_\mathcal{F} \in \mathcal{S}$ is sufficient for the first constraint in Eq.~\eqref{eq:moment_const} to be true, since $\mathbb{E}[\mathbf{\mu}_\mathcal{F}]$ is a convex combination of the instantiations of $\mathbf{\mu}_\mathcal{F}$ and $\mathcal{S}$ is a convex set. We rewrite the inner maximization problem as Eq.~\eqref{eq:inner_max}: 
\vspace{-0.1in}
 \begin{subequations}\label{eq:inner_max}
	\begin{align}
	& \sup_{\mathcal{F} \sim \Pi } & & \int_{\mathcal{S}}^{}{\big[\mathbf{x}^T \mathbf{Q}(\mathbf{\mu}_{\mathcal{F}}, \hat{\mathbf{\Sigma}}) \mathbf{x} + 2 \mathbf{x}^T \mathbf{b}(\mathbf{\mu}_{\mathcal{F}})  \big] d\mathcal{F}(\mathbf{\mu}_\mathcal{F}) }\\
	&s.t.     &    &  \int_{\mathcal{S}}^{}{d\mathcal{F}(\mathbf{\mu}_\mathcal{F})} = 1 \label{a:1}  \\
	&         &    &  \int_{\mathcal{S}}^{}{\big[(\mathbf{\mu}_\mathcal{F} - \hat{\mathbf{\mu}}) (\mathbf{\mu}_\mathcal{F} - \hat{\mathbf{\mu}})^T  \big]d\mathcal{F}(\mathbf{\mu}_\mathcal{F})} \preceq \gamma_2 \hat{\mathbf{\Sigma}} \label{a:2} \\
	&         &    &   \mathbf{\mu}_{\mathcal{F}} \in \mathcal{S} \label{a:3}, \forall \mathbf{\mu}_\mathcal{F}.
	\end{align}
\end{subequations}
The constraint Eq.\eqref{a:1} ensures that $\mathcal{F}$ is a valid probability distribution. 
The constraints Eq.\eqref{a:2} guarantee that $\mathcal{F}$ is in $\Pi$.
The constraint Eq.~\eqref{a:3} ensures that any random variable $\mathbf{\mu}_\mathcal{F} \sim \mathcal{F}$ must reside in $\mathcal{S}$.
Consequently, this constraint is actually an infinite dimensional constraint on the optimizer $\mathcal{F}$.
Later we introduce a technique called \textit{S-Lemma} to convert it to a finite dimensional constraint.
We derive the lagrange function of Eq.~\eqref{eq:inner_max}, where we temporily omit constraint Eq.~\eqref{a:3}, and pull the terms that are independent of $\mathcal{F}$ out of the integral:
\vspace{-0.1in}
\begin{equation*}
\begin{aligned}
& l(\mathcal{F} ,  t,  \mathbf{K}) = \bigg[ t + Tr\bigg( \big[ \gamma_2\hat{\mathbf{\Sigma}} + \hat{\mathbf{\mu}} \hat{\mathbf{\mu}}^T  \big]\mathbf{K} \bigg) \bigg] + \\
							& \int_{\mathcal{S}}^{}{ \bigg[  \underbrace{\mathbf{x}^T \mathbf{Q}(\mathbf{\mu}_{\mathcal{F}}, \hat{\mathbf{\Sigma}}) \mathbf{x} +  2 \mathbf{x}^T \mathbf{b}(\mathbf{\mu}_{\mathcal{F}}) - t - \mathbf{\mu}_{\mathcal{F}}^T \mathbf{K} \mathbf{\mu}_{\mathcal{F}}}_{:=f(\mathbf{\mu}_\mathcal{F})}     \bigg]  },
\end{aligned}
\end{equation*}
where $t \in \mathbb{R}$, and $\mathbf{K}$ is a real symmetric positive semi-definite matrix, and $Tr(\mathbf{X})$ returns  the trace of the matrix $\mathbf{X}$.
where  $f(\mathbf{\mu}_\mathcal{F}) \le 0, \forall \mathbf{\mu}_{\mathcal{F}} \in \mathcal{S}$ holds, since otherwise the solution to Eq.\eqref{eq:inner_max} is unbounded.

By duality~\citep{shapiro2001duality,delage2010distributionally,cheng2014distributionally}, the dual problem of  Eq.~\eqref{eq:inner_max} is formulated as the following minimization problem:
\vspace{-0.1in}
 \begin{equation}\label{eq:inner_max_dual}
\begin{aligned}
& \min_{t, \mathbf{K}} & &  t + Tr\bigg( \big[ \gamma_2\hat{\mathbf{\Sigma}} + \hat{\mathbf{\mu}} \hat{\mathbf{\mu}}^T  \big]\mathbf{K} \bigg) \\
&s.t.     &    &    f(\mathbf{\mu}_\mathcal{F})  \le 0, \forall \mathbf{\mu}_{\mathcal{F}} \in \mathcal{S} \\
&         &    &      t \in \mathbb{R}, \mathbf{K} \in \mathbb{S}^N_+
\end{aligned}
\end{equation}
where $\mathbb{S}^N_+$ is the positive semi-definite cone.  Strong duality holds between Eq.~\eqref{eq:inner_max_dual} and the original inner maximization problem. This is because for any $\gamma_1 > 0$ and $\gamma_2 > 0$, the estimated distribution $\hat{\mathcal{P}}$ is always in the relative interior of $\Pi$. Consequently, by Proposition 3.4 in \citet{shapiro2001duality} strong duality holds. Since Eq.~\eqref{eq:inner_max_dual} is a minimization problem, we can jointly minimize it with the outer minimization over $\mathbf{x}$, which results in the following:
\vspace{-0.1in}
 \begin{subequations}\label{eq:intermediate_MINT_DRO}
\begin{align}
& \min_{\mathbf{x}, t, \mathbf{K}} & &  t  + Tr\bigg( \big[ \gamma_2\hat{\mathbf{\Sigma}} + \hat{\mathbf{\mu}} \hat{\mathbf{\mu}}^T  \big]\mathbf{K} \bigg) \\
&s.t.     &    &    (\mathbf{\mu}_\mathcal{F} - \hat{\mathbf{\mu}})^T \hat{\mathbf{\Sigma}}^{-1} (\mathbf{\mu}_\mathcal{F} - \hat{\mathbf{\mu}}) \le \gamma_1, \forall \mathbf{\mu}_\mathcal{F} \in \mathcal{S} \label{b:1} \\
&         &    &    f(\mathbf{\mu}_\mathcal{F})  \le 0, \forall \mathbf{\mu}_{\mathcal{F}} \in \mathcal{S} \label{b:2} \\
&         &    &      t \in \mathbb{R}, \mathbf{K} \in \mathbb{S}^N_+
\end{align}
\end{subequations}
where constraint Eq.~\eqref{b:1} is equivalent to $\mathbf{\mu}_\mathcal{F} \in \mathcal{S}, \forall \mathbf{\mu}_\mathcal{F}$. We write it this way in order to emphasize its quadratic form. Constraints Eq.\eqref{b:1} and \eqref{b:2} are  infinite dimensional constraints. We apply a technique called \textit{S-Lemma} to transform them to finite dimensional constraints.  We first introduce the \textit{S-Lemma}:

\begin{lemma}[S-Lemma~\cite{boyd2004convex}]
Let $\mathbf{A}_1, \mathbf{A}_2 \in \mathbb{S}^n$, $\mathbf{b}_1, \mathbf{b}_2 \in \mathbb{R}^n$, $c_1, c_2 \in \mathbb{R}$,  where $\mathbb{S}^n$ is the subspace of symmetrix matrices in $\mathbb{R}^{n \times n}$. Suppose there exists an $\hat{\mathbf{x}} \in \mathbb{R}^n$ such that: $\hat{\mathbf{x}}^T \mathbf{A}_1 \hat{\mathbf{x}}  + 2 \mathbf{b}_1^T \hat{\mathbf{x}} + c_1 < 0$.
Then the following implication holds for any $\mathbf{x} \in \mathbb{R}^n$:
\begin{equation*}
\mathbf{x}^T \mathbf{A}_1 \mathbf{x}  + 2 \mathbf{b}_1^T \mathbf{x} + c_1 \le 0 \implies \mathbf{x}^T \mathbf{A}_2 \mathbf{x}  + 2 \mathbf{b}_2^T \mathbf{x} + c_2 \le 0
\end{equation*}
if and only if,  $\exists \lambda \ge 0: \begin{bmatrix}
\mathbf{A}_2 & \mathbf{b}_2 \\ \mathbf{b}_2^T & c_2
\end{bmatrix}
\preceq
\lambda \begin{bmatrix}
\mathbf{A}_1 & \mathbf{b}_1 \\ \mathbf{b}_1^T & c_1
\end{bmatrix}$.
\end{lemma}
Note that \textit{S-Lemma} only requires $\mathbf{A}_1$ and $\mathbf{A}_2$ to be real symmetric matrices. In order to apply \textit{S-Lemma} we need to have two quadratic functions. Constraint Eq.~\eqref{b:1} is a quadratic function in $\mathbf{\mu}_\mathcal{F}$. Thus, what remains is to convert Eq.~\eqref{b:2} to a quadratic function in $\mathbf{\mu}_\mathcal{F}$. Recall that the term,  $\mathbf{x}^T \mathbf{Q}(\mathbf{\mu}_{\mathcal{F}}, \hat{\mathbf{\Sigma}}) \mathbf{x} +  2 \mathbf{x}^T \mathbf{b}(\mathbf{\mu}_{\mathcal{F}})$ in $f(\mathbf{\mu}_\mathcal{F})$,  
is implicitly a quadratic function of $\mathbf{\mu}_\mathcal{F}$. We re-formulate  $\mathbf{Q}$ and $\mathbf{b}$ according to $\mathbf{\mu}_\mathcal{F}$, which results in Eq.\eqref{eq:quadratic_transform} (see the Appendix for details about this reformulation):
\vspace{-0.1in}
\begin{equation}\label{eq:quadratic_transform}
\begin{aligned}
& \mathbf{x}^T \mathbf{Q}(\mathbf{\mu}_{\mathcal{F}}, \hat{\mathbf{\Sigma}}) \mathbf{x} +  2 \mathbf{x}^T \mathbf{b}(\mathbf{\mu}_{\mathcal{F}}) = \\
\mathbf{\mu}_{\mathcal{F}} & \Bigg[ - (\alpha_2 + \alpha_3) \bigg(\mathbf{A}\odot \big(\mathbf{x}\mathbf{x}^T \big) \bigg)  \Bigg] \mathbf{\mu}_{\mathcal{F}}^T + \\
\mathbf{\mu}_\mathcal{F} ^T &\Bigg[ (\alpha_3+2\alpha_2)   diag(\mathbf{x})\cdot \mathbf{A} \cdot \mathbf{x}  - \alpha_1\mathbf{x} \Bigg] - \\
&\Bigg[ \alpha_1 \mathbf{1}^T\mathbf{x} - \mathbf{x}^T\bigg(  (\alpha_2+\alpha_3)\big( \mathbf{A}\odot \hat{\mathbf{\Sigma}} \big) + \alpha_2 \mathbf{A}   \bigg) \mathbf{x}  \Bigg],
\end{aligned}
\end{equation}
where $diag(\mathbf{x})$ returns a diagonal matrix with diagonal elements equal to $\mathbf{x}$. We substitute Eq.~\eqref{eq:quadratic_transform} back to $f(\mathbf{\mu}_\mathcal{F})$, which results in the following equivalence:
\vspace{-0.1in}
\begin{equation*}
\forall \mathbf{\mu}_\mathcal{F} \in \mathcal{S}: f(\mathbf{\mu}_\mathcal{F})  \le 0 \iff \mathbf{\mu}_{\mathcal{F}}^T \mathbf{R} \mathbf{\mu}_{\mathcal{F}} + \mathbf{\mu}_{\mathcal{F}}^T\mathbf{r} + z \le 0
\end{equation*}
where:
\vspace{-0.1in}
\begin{equation*}
\begin{aligned}
& \mathbf{R}  =  - (\alpha_2 + \alpha_3) \bigg(\mathbf{A}\odot \big(\mathbf{x}\mathbf{x}^T \big) \bigg) - \mathbf{K} \\
& \mathbf{r}  = (\alpha_3+2\alpha_2)diag(\mathbf{x})\cdot \mathbf{A} \cdot \mathbf{x}  - \alpha_1 \mathbf{x}   \\
& z           = \alpha_1 \mathbf{1}^T\mathbf{x} - \mathbf{x}^T\bigg(  (\alpha_2+\alpha_3)\big( \mathbf{A}\odot \hat{\mathbf{\Sigma}} \big) + \alpha_2 \mathbf{A}   \bigg) \mathbf{x} - t,
\end{aligned}
\end{equation*}
which results in a compact form of $f(\mathbf{\mu}_\mathcal{F})$:
\vspace{-0.1in}
\begin{equation*}
f(\mathbf{\mu}_\mathcal{F}) = \mathbf{\mu}_{\mathcal{F}}^T \mathbf{R} \mathbf{\mu}_{\mathcal{F}} + \mathbf{\mu}_{\mathcal{F}}^T \mathbf{r} + z
\end{equation*}
Note that for any $\gamma_1 >0$ the inequality in constraint Eq.~\eqref{b:1} is strict when $\mathbf{\mu}_\mathcal{F}=\hat{\mathbf{\mu}}$. Consequently, by \textit{S-Lemma}, for any $\mathbf{\mu}_{\mathcal{F}} \in \mathcal{S}$ the implication, $Eq.~\eqref{b:1} \implies \mathbf{\mu}_{\mathcal{F}}^T \mathbf{R} \mathbf{\mu}_{\mathcal{F}} + \mathbf{\mu}_{\mathcal{F}}^T\mathbf{r} + z$, 
is equivalent to Eq.\eqref{eq:linear_matrix_ine}:
\vspace{-0.1in}
\begin{equation}\label{eq:linear_matrix_ine}
\exists \lambda \ge 0:
\begin{bmatrix} \mathbf{R} & \frac{1}{2}\mathbf{r} \\ \frac{1}{2}\mathbf{r}^T & z \end{bmatrix} \preceq \lambda \begin{bmatrix} \hat{\mathbf{\Sigma}}^{-1} & -\hat{\mathbf{\Sigma}}^{-1}\hat{\mathbf{\mu}} \\ -\hat{\mathbf{\mu}}^T \hat{\mathbf{\Sigma}}^{-1} & \big( \hat{\mathbf{\mu}}^T \hat{\mathbf{\Sigma}}^{-1} \hat{\mathbf{\mu}} - \gamma_1 \big).  \end{bmatrix}
\end{equation}

The two infinite dimensional constraints Eq.\eqref{b:1} and \eqref{b:2} are thereby converted into a finite dimensional constraint Eq.~\eqref{eq:linear_matrix_ine}.
Additionally, the objective function in Eq.~\eqref{eq:intermediate_MINT_DRO} is linear in its optimizer.

The last issue is that we still have two sources of non-convexity in Eq.~\eqref{eq:intermediate_MINT_DRO}: first, $\mathbf{x}$ is binary, and second, the constraint
represented by Eq.~\eqref{eq:linear_matrix_ine} is not convex in $\mathbf{x}$ because of three terms involving in $\mathbf{R}$, $\mathbf{r}$ and $z$:
\vspace{-0.1in}
\begin{equation}\label{eq:nonconvex_term}
\mathbf{x}\mathbf{x}^T, \,\, diag(\mathbf{x})\mathbf{A}\mathbf{x}, \,\, \mathbf{x}^T \bigg(   (\alpha_2+\alpha_3)(\mathbf{A}\odot \hat{\mathbf{\Sigma}}) + \alpha_2\mathbf{A} \bigg) \mathbf{x}.
\end{equation}
To deal with the first issues, we relax the feasible region of $\mathbf{x}$ to $[-1, 1]^N$.
To address the second, we next apply SDP relaxation to transform Eq.~\eqref{eq:intermediate_MINT_DRO} into a convex optimization problem.



First, let us introduce a matrix $\mathbf{X} = \mathbf{x}\mathbf{x}^T$.
Then the following three relationships hold (see the Appendix for detailed proof):
\vspace{-0.1in}
\begin{equation}\label{eq:trace_relation}
\begin{aligned}
& (r1): \bigg(\mathbf{A} \odot \big( \mathbf{x}\mathbf{x}^T \big) \bigg) =   \bigg(\mathbf{A} \odot \mathbf{X}  \bigg) \\
& (r2):  diag(\mathbf{x})\cdot \mathbf{A} \cdot \mathbf{x}  =  diag(\mathbf{A}\mathbf{X}) \\
& (r3): \mathbf{x}^T \bigg(   (\alpha_2+\alpha_3)(\mathbf{A}\odot \hat{\mathbf{\Sigma}}) + \alpha_2\mathbf{A} \bigg) \mathbf{x}   = \\
& \qquad \,  (\alpha_2+\alpha_3)Tr\big((\mathbf{A}\odot\hat{\mathbf{\Sigma}})\mathbf{X} \big)  + \alpha_2Tr\big(\mathbf{A}\mathbf{X} \big)
\end{aligned}
\end{equation}

One problem is that the feasible regions involving $\mathbf{X}$ and $\mathbf{x}$ are nonconvex because of the equality $\mathbf{X}=\mathbf{x}\mathbf{x}^T$. In order to transform the feasible regions to be convex, we apply a two-step relaxation. The first step is to relax the equality and enforce the diagonal elements of $\mathbf{X}$ equal to one, which results in: $\mathbf{X} \succeq \mathbf{x}\mathbf{x}^T$ and $\mathbf{X}_{ii} = 1, \forall i=1,\cdots, N$.
This step transforms the feasible region of $\mathbf{X}$ to a positive semi-definite cone, which is a convex set. However, we still have a nonconvex term $\mathbf{x}\mathbf{x}^T$. To handle this, in the second step we apply \textit{Schur Complement} to transform $\mathbf{X} \succeq \mathbf{x}\mathbf{x}^T$ to the  linear matrix inequality: $\begin{bmatrix}\mathbf{X} & \mathbf{x} \\ \mathbf{x}^T & 1 \end{bmatrix} \succeq 0$.
Combining the relationships in Eq.~\eqref{eq:trace_relation} with the results of the two-step relaxation above,  the three nonconvex terms in Eq.~\eqref{eq:nonconvex_term} can be represented as the following convex set:
\vspace{-0.1in}
\begin{equation*}
\begin{aligned}
& \hat{\mathbf{R}}  =  -(\alpha_2+\alpha_3)\big(\mathbf{A} \odot \mathbf{X} \big)  - \mathbf{K} \\
& \hat{\mathbf{r}}  = (2\alpha_2 + \alpha_3) \underbrace{diag(\mathbf{A}\mathbf{X})}_{(\ast)} - \alpha_1\mathbf{x}  \\
& \hat{z}           = \alpha_1 \mathbf{1}^T \mathbf{x} - \bigg(  (\alpha_2+\alpha_3)Tr\big( (\mathbf{A}\odot\hat{\mathbf{\Sigma}})\mathbf{X} \big) + \alpha_2 Tr\big(\mathbf{A}\mathbf{X}\big)  \bigg) \\
& \qquad - t \\
& \begin{bmatrix}\mathbf{X} & \mathbf{x} \\ \mathbf{x}^T & 1 \end{bmatrix} \succeq 0 , \mathbf{X}_{ii} = 1, \forall i=1,\cdots, N.
\end{aligned}
\end{equation*}

With a slight abuse of notation, the operator $diag(\mathbf{A}\mathbf{X})$ in $(\ast)$ extracts the diagonal elements of $\mathbf{A}\mathbf{X}$ as a column vector. Finally, by substituting $\hat{\mathbf{R}}$, $\hat{\mathbf{r}}$ and $\hat{z}$ to the corresponding matrices in Eq.~\eqref{eq:linear_matrix_ine} we obtain the following Semidefinite Program which approximately solves \texttt{MINT\_DRO} (after we project the optimal solution $\mathbf{x}$ of this problem into $\{0,1\}^N$, for example, by rounding):
\vspace{-0.2in}
 \begin{equation}\label{eq:final_SDP}
\begin{aligned}
& \min_{\mathbf{x}, \mathbf{X}, t, \mathbf{K}, \lambda} & &  t + Tr\bigg( \big[ \gamma_2\hat{\mathbf{\Sigma}} + \hat{\mathbf{\mu}} \hat{\mathbf{\mu}}^T  \big]\mathbf{K} \bigg) \\
&s.t.     &    &    \begin{bmatrix} \hat{\mathbf{R}} & \frac{1}{2}\hat{\mathbf{r}} \\ \frac{1}{2}\hat{\mathbf{r}}^T & \hat{z} \end{bmatrix} \preceq \lambda \begin{bmatrix} \hat{\mathbf{\Sigma}}^{-1} & -\hat{\mathbf{\Sigma}}^{-1}\hat{\mathbf{\mu}} \\ -\hat{\mathbf{\mu}}^T \hat{\mathbf{\Sigma}}^{-1} & \big( \hat{\mathbf{\mu}}^T \hat{\mathbf{\Sigma}}^{-1} \hat{\mathbf{\mu}} - \gamma_1 \big)  \end{bmatrix} \\
&         &    &       \begin{bmatrix}\mathbf{X} & \mathbf{x} \\ \mathbf{x}^T & 1 \end{bmatrix} \succeq 0 , \mathbf{X}_{ii} = 1, \forall i=1,\cdots, N \\
&         &    &    \mathbf{x} \in [-1, 1]^N,  t \in \mathbb{R},  \mathbf{K} \in \mathbb{S}^N_+, \mathbf{X} \in \mathbb{S}^N_+, \lambda \ge 0
\end{aligned}
\end{equation}

\section{Theoretical Analysis}\label{S:th}
In this section we present a probabilistic argument that the
uncertainty set $\Pi$ defined in Eq.~\eqref{eq:uncertainty_set}
contains the ground-truth distribution $\mathcal{P}$ with high
probability.
This, in turn, implies that with high probability our model
\texttt{MINT\_DRO} is robust with respect to the \emph{unknown} ground-truth distribution.

We show that the ground-truth distribution $\mathcal{P}$ belongs to
$\Pi$ with high probability in two steps, arguing first that (C1) and,
subsequently, that (C2) below hold with high probability, where (C1)
and (C2) are defined as follows:
\vspace{-0.1in}
\[
(\mathbb{E}_{\mathbf{\pi} \sim \mathcal{P}}[\mathbf{\pi}] - \hat{\mathbf{\mu}})^T \hat{\mathbf{\Sigma}}^{-1} (\mathbb{E}_{\mathbf{\pi} \sim \mathcal{P}}[\mathbf{\pi}] - \hat{\mathbf{\mu}}) \le \gamma_1 \\ \tag{C1} 
\]
\[
\mathbb{E}_{\mathbf{\pi} \sim \mathcal{P}}\big[(\mathbf{\pi} - \hat{\mathbf{\mu}})(\mathbf{\pi} - \hat{\mathbf{\mu}})^T \big] \preceq \gamma_2 \hat{\mathbf{\Sigma}} \tag{C2} 
\]

The arguments in the first step are  based on Lemma~\ref{th:bound_mean}. For space limitation we defer its proof to the appendix.
\begin{lemma}\label{th:bound_mean}
Let $\mathbf{\mu}$ and $\mathbf{\Sigma}$ denote the mean and
covariance matrix of the ground-truth distribution $\mathcal{P}$, and
suppose that $\hat{\mathbf{\mu}}$ is estimated from $M$ samples, $\hat{\mathbf{\mu}}=\frac{1}{M}\sum_{i=1}^{M}{\zeta_i}$, where $\zeta_i$  is bounded: ${\Vert \mathbf{\Sigma}^{-1/2}(\zeta_i - \mathbf{\mu}) \Vert}_2^2 \le R^2, \forall i$.
Then $\hat{\mathbf{\mu}}$ satisfies the following constraint with probability at least $1-\delta_1$:
\vspace{-0.1in}
\[
(\mathbf{\mu} - \hat{\mathbf{\mu}})^T \mathbf{\Sigma}^{-1} (\mathbf{\mu} - \hat{\mathbf{\mu}}) \le \beta(\delta),
\]
where $\beta(\delta_1)=\frac{R^2}{M}\bigg( 2 + \sqrt{2\log{\frac{1}{\delta_1}}} \bigg)^2$.
\end{lemma}
\vspace{-0.1in}
We assume the estimated covariance matrix $\hat{\mathbf{\Sigma}}$ is
close  to $\mathbf{\Sigma}$. Then, if we let $\gamma_1 >
\beta(\delta_1)$ and note that $\mathbf{\mu} =
\mathbb{E}_{\mathbf{\pi} \sim \mathcal{P}}[\mathbf{\pi}]$, a direct
application of  Lemma~\ref{th:bound_mean} implies that (C1) holds with probability at least $1-\delta_1$.

The arguments in the second step rely on the result due to~\cite{delage2010distributionally}:
\begin{lemma}[\citet{delage2010distributionally}]\label{th:bound_cov}
Suppose that $\zeta_i$ is distributed according to $\mathcal{G}$, and the mean $\mathbf{\mu}$ of the distribution is known and used to formulate the estimated covariance matrix $\hat{\mathbf{\Sigma}}$, which  is estimated from $M$ samples: $\hat{\mathbf{\Sigma}}=(1/M)\sum_{i=1}^{M}{\big(\zeta_i - \mathbf{\mu}\big) \big( \zeta_i - \mathbf{\mu} \big)^T}$, where $\zeta_i$ is bounded: ${\Vert \mathbf{\Sigma}^{-1/2}(\zeta_i - \mathbf{\mu}) \Vert}_2^2 \le R^2, \forall i$.
Then with probability at least $1 - \delta_2$: 
\vspace{-0.1in}
\[
\mathbf{\Sigma} \preceq \frac{1}{1 - \alpha(\delta_2)} \hat{\mathbf{\Sigma}},
\]
where $\alpha(\delta_2)=(R^2 / \sqrt{M})\bigg( \sqrt{1 - N/R^4} + \sqrt{\log{1 / \delta_2}} \bigg)$, $M > R^4\bigg( \sqrt{1 - N / R^4} +  \sqrt{\log{1/\delta_2}} \bigg)^2$ and $N$ is the dimensions of $\mathbf{\mu}$.
\end{lemma}
\vspace{-0.1in}
In order to use Lemma~\ref{th:bound_cov} we assume that the estimated
mean $\hat{\mathbf{\mu}}$ is close to the ground-truth $\mathbf{\mu}$.
Given this assumption, showing that (C2) holds with high probability
is equivalent to show that the following holds with high probability:
\vspace{-0.1in}
\[ 
\mathbb{E}_{\mathbf{\pi} \sim \mathcal{P}}[\mathbf{\pi} \mathbf{\pi}^T] \preceq \gamma_2 \hat{\mathbf{\Sigma}} + \mathbf{\mu}\mathbf{\mu}^T
\]
by Lemma~\ref{th:bound_cov}, the above is true with high probability when: $\frac{1}{1 - \alpha(\delta_2)} \hat{\mathbf{\Sigma}} \preceq \gamma_2 \hat{\mathbf{\Sigma}} + \mathbf{\mu}\mathbf{\mu}^T$.
Consequently, by setting $\gamma_2 > \frac{1}{1 - \alpha(\delta_2)}$, such that the effects of $\mathbf{\mu}\mathbf{\mu}^T$ are negligible, we conclude that (C2) holds with probability at least $1-\delta_2$.

Finally, by a union bound we obtain probabilistic guarantees that the
uncertainty set $\Pi$ contains $\mathcal{P}$. 
\begin{theorem}\label{th:prob_guarantee}
With probability at least $1-\delta$, where $\delta=\delta_1+\delta_2$,  the uncertainty set $\Pi$ defined in Eq.~\eqref{eq:uncertainty_set} contains the ground-truth distribution $\mathcal{P}$.
\end{theorem}
\vspace{-0.1in}
\begin{proof}
The detailed proof is deferred to the appendix.
\end{proof}
\vspace{-0.1in}

We now demonstrate how to utilize the probabilistic arguments to make
sensible choice for $\gamma_1$. The value of $\gamma_2$ can be
similarly obtained. Note that $\gamma_1 > \beta(\delta_1)$ is
necessary for (C1) to hold. Consider a network with $N=128$
nodes. Assume $\mathbf{\Sigma}$ is diagonal with diagonal elements
equal to $0.01$, which is reasonable when a single estimator is used
to estimate $\mathcal{P}$ and the maliciousness probabilities of nodes
are independent. A reasonable estimate of $R$ is $\sqrt{128 \times
  2}$, which is the radius of the circumcircle sphere of a hypercube
with length of side equal to one. If $M=5$ and $\delta_1=0.05$, then
$\beta(0.05)=1012$. Therefore in order for $\Pi$ to contain
$\mathcal{P}$ with probability $\ge 0.95$, we need $\gamma_1 \ge
1012$. Similarly, for a network with $N=500$ nodes, we want $\gamma_1 \ge 3956$.

\vspace{-0.1in}

\section{Experiments}\label{S:exp}
In this section we present experimental results to show the effectiveness of our approach. Our experiments were conducted on both synthetic and real-world network structures, although in all cases the distribution $\mathcal{P}$ over maliciousness of nodes was derived using real data. We considered two types of network generative models to construct synthetic networks: Barabasi-Albert (BA)~\cite{barabasi1999emergence} and Watts-Strogatz networks (Small-World)~\cite{watts1998collective}. BA is characterized by its power-law degree distribution, where the probability that a randomly selected node has $k$ neighbors is proportional to $k^{-r}$. For the BA model we experimented with three variants, \textit{BA-1}, \textit{BA-2}, and \textit{BA-3}, which differ in the value of the exponent $r$ of their power-law degree distributions. For Small-World networks we also experimented with three variants, \textit{SW-1}, \textit{SW-2}, and \textit{SW-3}, that have different local clustering coefficients. For both networks we generated instances with $N=128$ nodes. For real-world networks, we used a network extracted from Facebook data~\cite{leskovec2012learning} which consisted of $4039$ nodes and $88234$ edges. 
We experimented with randomly sampled sub-networks with $N=500$ nodes. For space limitation the statistics of the networks used in our experiments are listed in the appendix.


For fair comparison with MINT (the state-of-the-art alternative), we used the same experimental setup as \citet{yu2018removing}. In all of our experiments, we derived the ground-truth distribution $\mathcal{P}$ as follow. We start with a dataset $\mathbf{D}$ which includes malicious and benign instances (the meaning of these designations is domain specific). The dataset $\mathbf{D}$ is partitioned into three subsets: $\mathbf{D}_{train}$, $\mathbf{D}_1$ and $\mathbf{D}_2$, with the ratio of $0.3:0.6:0.1$. Our first step is to learn a probabilistic predictor of maliciousness as a function of a feature vector $\mathbf{x}$, $\hat{p}(\mathbf{x})$, on $\mathbf{D}_{train}$. Then we randomly assign malicious and benign feature vectors from $\mathbf{D}_2$ to the nodes on the network, assigning $10\%$ of nodes with malicious features and $90\%$ with benign feature vectors. For each node we use its assigned feature vector $\mathbf{x}$ to obtain our estimated probability of this node being malicious, $\hat{p}(\mathbf{x})$; This gives us the estimated maliciousness probability distribution $\hat{\mathcal{P}}$. This is the distribution used to solve the model \texttt{MINT}, and also the distribution used to construct the uncertainty set $\Pi$ in our model. To ensure that our evaluation reasonably reflects realistic limitations of the knowledge about the ground-truth distribution $\mathcal{P}$, we train another predictor $p(\mathbf{x})$ usign $\mathbf{D}_{train} \bigcup \mathbf{D}_1$. Applying this new predictor to the nodes and their assigned feature vectors, we obtain a distribution $\mathcal{P}^\ast$ which we use to evaluate effectiveness. 

We conducted three sets of experiments. In the first set of experiments we used synthetic networks and used data from the Spam~\cite{cormack2008email} dataset 
To simulate estimation error of $\mathcal{P}$, we add white Gaussian noise to the evaluation distribution $\mathcal{P}^\ast$. The standard deviation of the noise is increased from $0.1$ to $0.5$ to simulate different magnitudes of the estimation error.

In the second set of experiments we used real-world networks from Facebook and used Hate Speech data~\cite{davidson2017automated} collected from Twitter to obtain $\mathcal{P}$ as discussed above. 
We categorized this dataset into two classes in terms of whether a tweet represents Hate Speech.
After categorization, the total number of tweets is $24783$, of which $1430$ are Hate Speech. 
We add white Gaussian noise to $\mathcal{P}^{\ast}$ to simulate estimation error as discussed above. Note that in this set of experiments we used \textit{real data} for both the networks and the maliciousness probabilities $\mathcal{P}$.

In the third set of experiments we considered the scenario that instead of being random, the location of the malicious nodes on the network is strategically determined. This scenario is not vacuous: in reality, for example, the nodes that have high degrees (e.g., celebrities with lots of followers on Twitter) may be targeted in order to maximize the influence of commercial advertisements~\cite{kempe2003maximizing}.
We conducted this set of experiments on  synthetic networks. A set of nodes is greedily selected from the network to maximize the number of unique neighbors connecting to them. Then we assign malicious feature vectors to these nodes. 

\begin{figure}[ht]
\centering
\setlength{\tabcolsep}{0.1pt}
\begin{tabular}{c}
\includegraphics[width=3.2in]{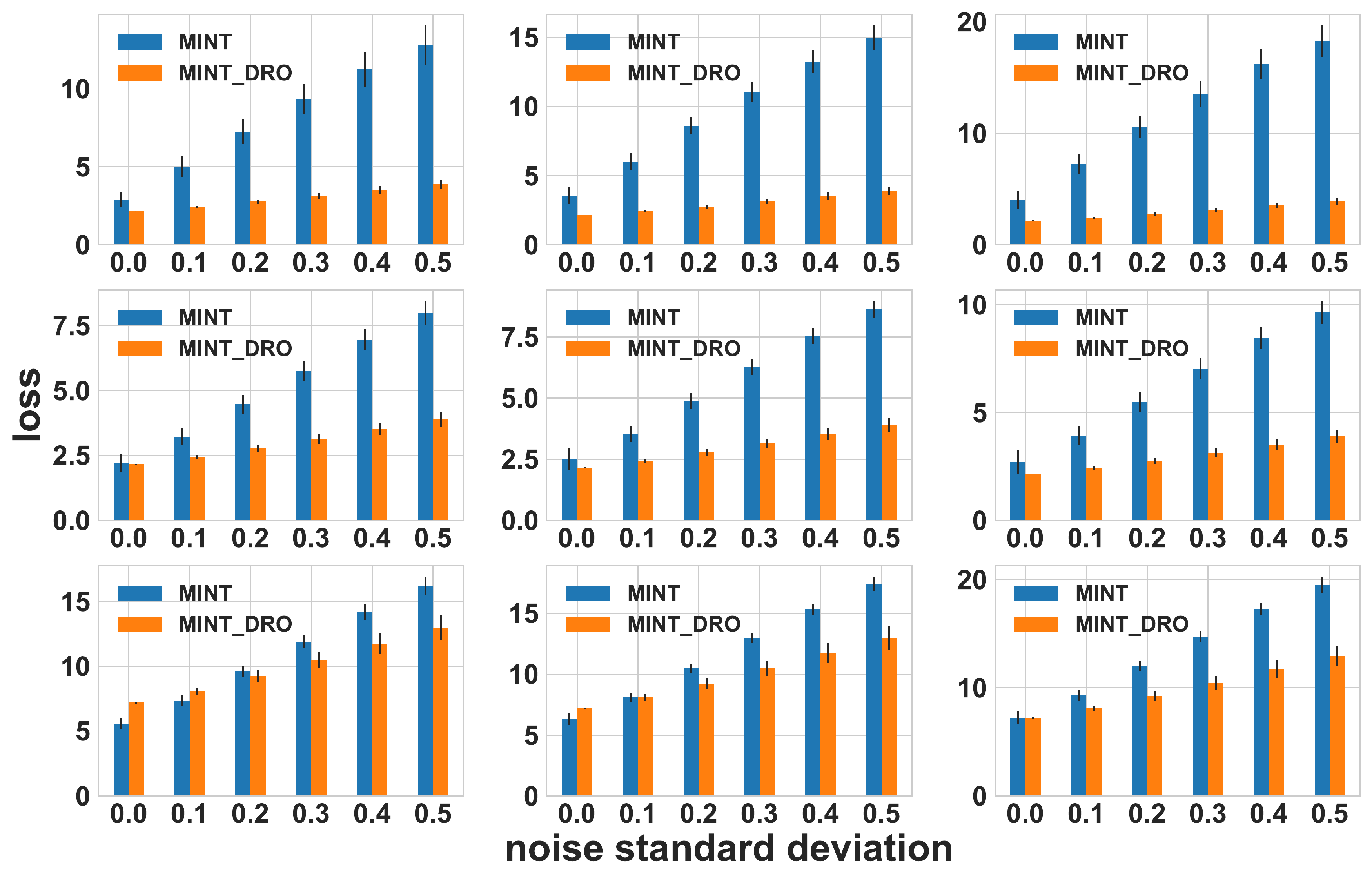}
\end{tabular} 
\vspace{-0.1in}
\caption{Experimental results on  BA networks. The three columns correspond to results on \textit{BA-1}, \textit{BA-2} and \textit{BA-3}, respectively. \textbf{Top row}: $(0.2, 0.7, 0.1)$; \textbf{Middle row}: $(0.7, 0.2, 0.1)$; \textbf{Bottom row}: $(\frac{1}{3}, \frac{1}{3}, \frac{1}{3})$.}
\label{fig:synthetic_BA}
\end{figure}

\begin{figure}[ht]
\centering
\setlength{\tabcolsep}{0.1pt}
\begin{tabular}{c}
\includegraphics[width=3.2in]{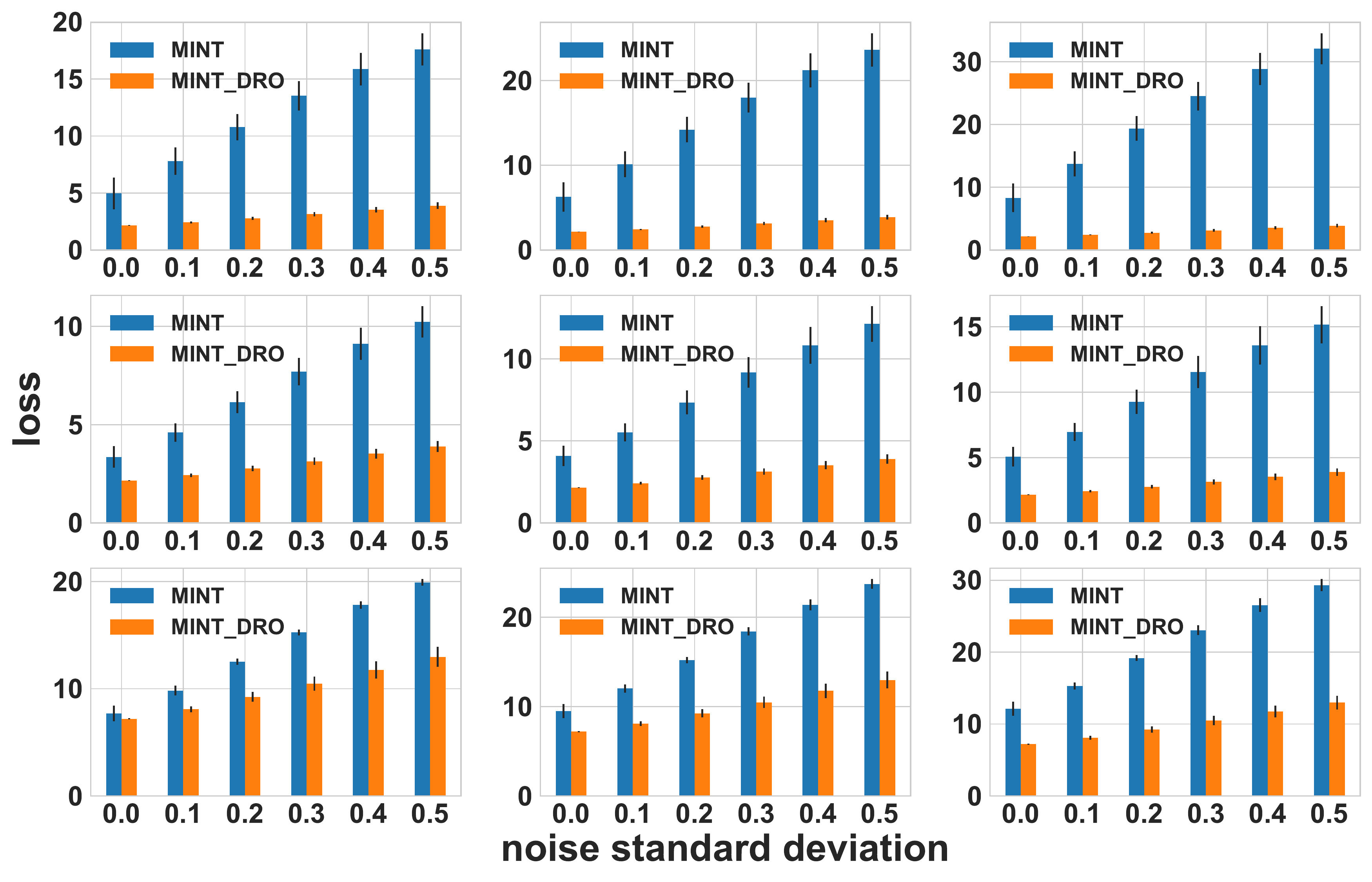} 
\end{tabular}
\vspace{-0.1in}
\caption{Experimental results on  Small-World networks. The three columns correspond to results on \textit{SW-1}, \textit{SW-2} and \textit{SW-3}, respectively. \textbf{Top row}: $(0.2, 0.7, 0.1)$; \textbf{Middle row}: $(0.7, 0.2, 0.1)$; \textbf{Bottom row}: $(\frac{1}{3}, \frac{1}{3}, \frac{1}{3})$.}
\label{fig:synthetic_SW}
\end{figure}

\paragraph{Experiment Results}
We compared our model with a state-of-the-art approach \texttt{MINT}. The average losses for our first set of experiments where $\mathcal{P}$ was simulated from Spam data are shown in Figures~\ref{fig:synthetic_BA} and \ref{fig:synthetic_SW}. 
The experimental results on BA are showed in Figure~\ref{fig:synthetic_BA}, with the three columns  corresponding to \textit{BA-1}, \textit{BA-2} and \textit{BA-3}, respectively. 
The experimental results on Small-World are shown in Figure~\ref{fig:synthetic_SW}, where the three columns correspond to \textit{SW-1}, \textit{SW-2}, and \textit{SW-3}. In both figures, each row corresponds to a combination of trade-off parameters $(\alpha_1, \alpha_2, \alpha_3)$; for example, $(0.2, 0.7, 0.1)$ corresponds to $(\alpha_1=0.2, \alpha_2=0.7, \alpha_3=0.1)$. 
Each bar was obtained by averaging over $30$ randomly generated network topologies. 

The experimental results indicate that on both BA and Small-World networks our model \texttt{MINT\_DRO} is significantly more robust than \texttt{MINT}. 
Additionally,  when no noise is added to the evaluation distribution $\mathcal{P}^{\ast}$ (left-most bars in all subplots), \texttt{MINT\_DRO} is  more robust than \texttt{MINT} except for a few cases. 
this  indicates that the \textit{generalization} ability of \texttt{MINT\_DRO} is better than \texttt{MINT}.

\vspace{-0.1in}

\vspace{-0.1in}

\begin{figure}[ht]
\centering
\setlength{\tabcolsep}{0.1pt}
\begin{tabular}{c}
\includegraphics[width=3.2in]{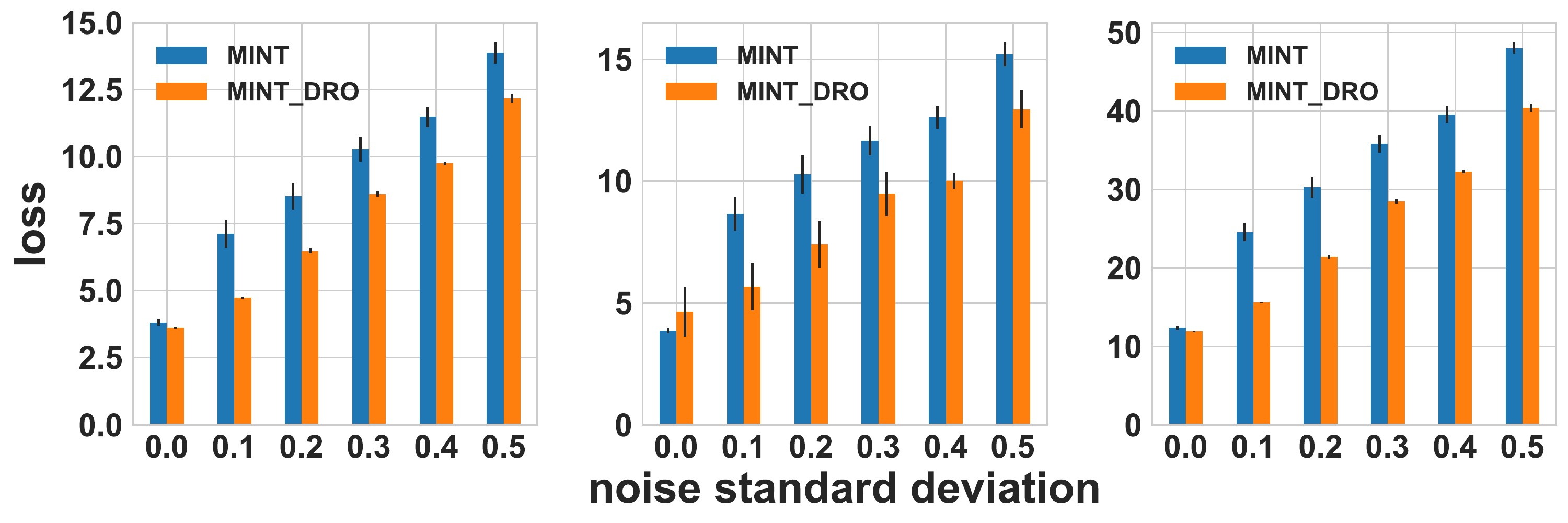} 
\end{tabular}
\vspace{-0.1in}
\caption{Experimental results on Facebook networks. The three columns correspond to $(0.2, 0.7, 0.1)$, $(0.7, 0.2, 0.1)$, and $(\frac{1}{3}, \frac{1}{3}, \frac{1}{3})$.}
\label{fig:facebook}
\end{figure}

The average loss on Facebook data is showed in Figure~\ref{fig:facebook}, with the three columns corresponding to $(0.2, 0.7, 0.1)$, $(0.7, 0.2, 0.1)$, and $(\frac{1}{3}, \frac{1}{3}, \frac{1}{3})$.  In this experiment, both the networks and the data used to simulate maliciousness probabilities are \textit{real} data. Each bar was averaged over $30$ randomly sampled networks. Our model \texttt{MITN\_DRO} is significantly more robust than \texttt{MINT} except for the cases where no noise is added. In this case, \texttt{MINT\_DRO} is only worse than \texttt{MINT} at the left-most bars in the middle figure, although the difference is not significant. This is actually expected since \texttt{MINT\_DRO} considers the \textit{worst-case} scenario, which results in a decision that may be slightly conservative in no noise setting.  One observation is that the Facebook networks used in this experiment are dramatically different from the simulated networks in terms of graph statistics (see the appendix for the detailed statistics). Particularly, the Facebook networks are disconnected, highly sparse, and have approximately $16\%$ nodes that have zero degree. Therefore the robustness exhibited in Figure~\ref{fig:facebook} provides strong evidence to the effectiveness of \texttt{MINT\_DRO}. 
\vspace{-0.1in}
\vspace{-0.1in}

\begin{figure}[ht]
\centering
\setlength{\tabcolsep}{0.1pt}
\begin{tabular}{c}
\includegraphics[width=3.3in]{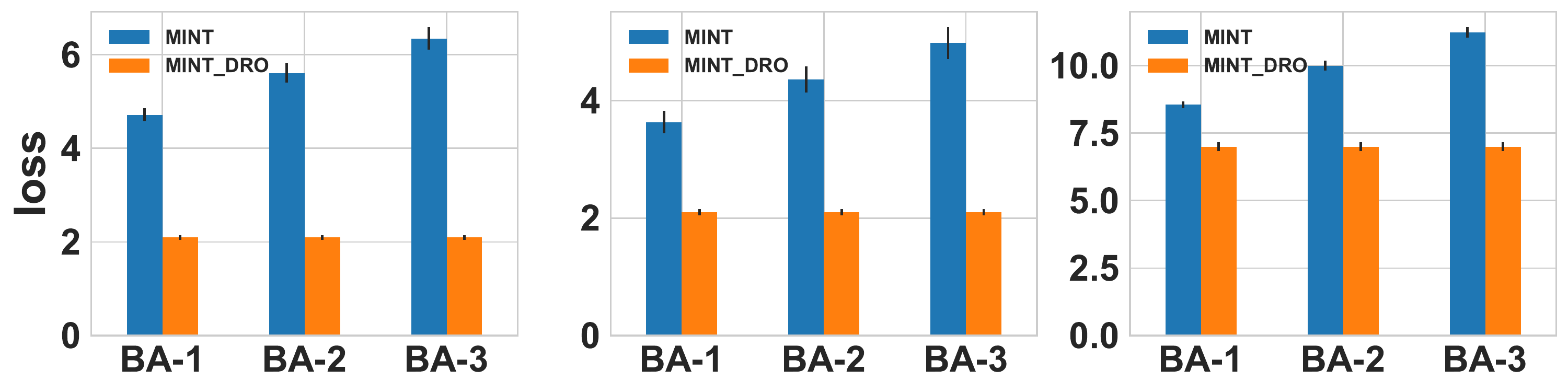}
\end{tabular}
\caption{Experimental results on the robustness to strategic selection of malicious nodes on BA networks. \textbf{Left}: $(0.2, 0.7, 0.1)$; \textbf{Middle}: $(0.7, 0.2, 0.1)$; \textbf{Right}: $(\frac{1}{3}, \frac{1}{3}, \frac{1}{3})$.}
\label{fig:placement_BA}
\end{figure}

\begin{figure}[ht]
\centering
\setlength{\tabcolsep}{0.1pt}
\begin{tabular}{c}
\includegraphics[width=3.3in]{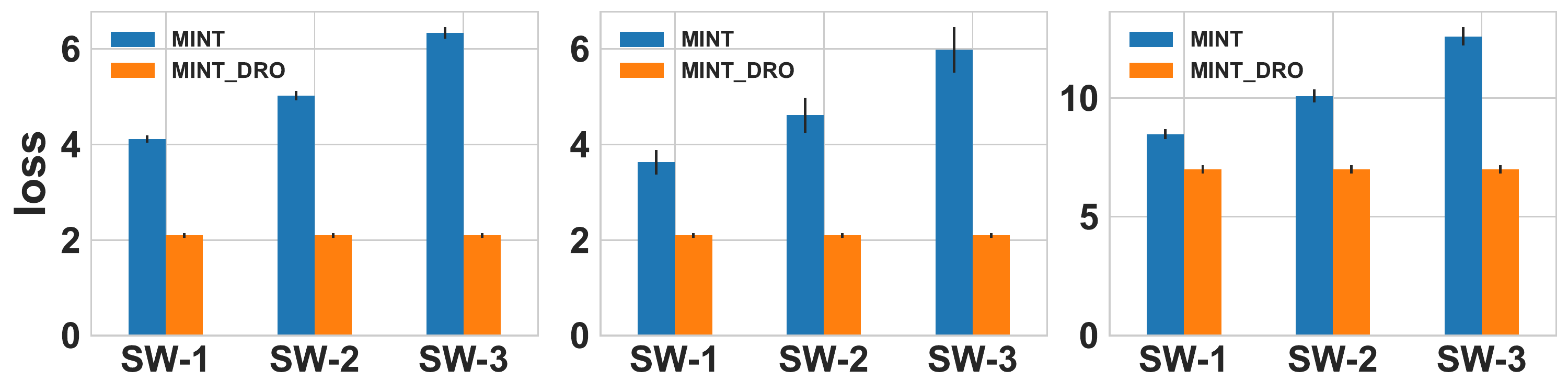} 
\end{tabular}
\caption{Experimental results on the robustness to strategic selection of malicious nodes on Small-World networks. \textbf{Left}: $(0.2, 0.7, 0.1)$; \textbf{Middle}: $(0.7, 0.2, 0.1)$; \textbf{Right}: $(\frac{1}{3}, \frac{1}{3}, \frac{1}{3})$}
\label{fig:placement_SW}
\end{figure}

The average loss on the third set of experiments are shown in Figures~\ref{fig:placement_BA} and \ref{fig:placement_SW} for BA and Small-World networks, respectively. For both figures the three columns correspond to  $(0.2, 0.7, 0.1)$, $(0.7, 0.2, 0.1)$, and $(\frac{1}{3}, \frac{1}{3}, \frac{1}{3})$. The results show that \texttt{MINT\_DRO} is more robust than \texttt{MINT} across all settings. Recall that the loss function of \texttt{MINT} and \texttt{MINT\_DRO} depends on the estimated covariance  matrix $\hat{\mathbf{\Sigma}}$, which encodes correlation information of the distribution $\hat{\mathcal{P}}$. 
When the actual maliciousness of nodes become correlated as we simulated in this experiment, the performance of \texttt{MINT} degrades since it is using the estimated distribution $\hat{\mathcal{P}}$ which now significantly deviates from the true distribution. 
When $\gamma_1$ and $\gamma_2$ are appropriately selected, $\Pi$  contains the distribution that characterizes the strategic correlation simulated in this experiment, resulting in significantly better robustness.

One may argue that instead of resulting from the robustness against correlation in the maliciousness distribution that comes from strategic decision about where to place the malicious nodes, the robustness exhibited in Figures~\ref{fig:placement_BA} and \ref{fig:placement_SW} stems solely from the fact that \texttt{MINT\_DRO} is more robust than \texttt{MINT} when no noise is added to $\mathcal{P}^{\ast}$.
However, consider the left-most bars in the lower-left subplot of Figure~\ref{fig:synthetic_BA}.
In this setting \texttt{MINT\_DRO} performs worse than \texttt{MINT}. 
Now, consider another setting where the experimetal setup is identical except that the malicious nodes are strategically chosen. 
This setting corresponds to the left-most bars in the right subplot of Figure~\ref{fig:placement_BA} where \texttt{MINT\_DRO} performs better than \texttt{MINT}.
Similar observations can be found on Small-World networks.
Consequently, we can see that a major advantage of \texttt{MINT\_DRO} is in its robustness even when the location of the malicious nodes on the graph is itself chosen strategically.

\vspace{-0.1in}
\vspace{-0.1in}
\vspace{-0.1in}

\section{Conclusion}
We considered the problem of removing malicious nodes from a network under uncertainty. We designed a model that considers the uncertainty around the estimated maliciousness probabilities, and makes decision under the \textit{worst-case} scenario. 
We then proposed a principled algorithmic
technique for solving it approximately based on duality combined with Semidefinite Programming relaxation. We theoretically proved that our model is robust with respect to the ground-truth, and experimentally showed that our model is more robust than the state of the art.

\bibliography{paper}
\bibliographystyle{icml2019}

\onecolumn
\section*{Appendix}
\setcounter{section}{0}
\section{Proof of Lemma~4.1}
The proof is a  generalization of a result proved by~\citeauthor{shawe2003estimating}. For completeness we list their result in Lemma~\ref{th:inequality}.
\begin{lemma}\label{th:inequality}
\cite{shawe2003estimating} \\
Assume $\zeta \in \mathbb{R}^N$ is a random variable satisfying:
\begin{equation*}
\begin{aligned}
& \mathbb{E}[\zeta] = \mathbf{0}\\
& \mathbb{E}[\zeta {\zeta}^T] = \mathbf{I} \\
& {\Vert \zeta \Vert}_2^2 \le R^2,
\end{aligned}
\end{equation*}
where the last inequality bounds the support of $\zeta$. Let $\{\zeta_i\}_{i=1}^{M}$ be a set of $M$ independently and ramdomly sampled instances of $\zeta$. Then with probability at least $(1 - \delta)$, the following inequality holds:
\begin{equation*}\label{eq:inequality}
\left\Vert \frac{1}{M}\sum_{i=1}^{M}{\zeta_i} \right\Vert^2 \le \frac{R^2}{M}\bigg( 2 + \sqrt{2\log{\frac{1}{\delta}}} \bigg)^2
\end{equation*}
\end{lemma}

In what follows we prove Lemma~\ref{lemma4.1}:
\begin{customlemma}{4.1}\label{lemma4.1}
Let $\mathbf{\mu}$ and $\mathbf{\Sigma}$ denote the mean and
covariance matrix of the ground-truth distribution $\mathcal{P}$, and
suppose that $\hat{\mathbf{\mu}}$ is estimated from $M$ samples, $\hat{\mathbf{\mu}}=\frac{1}{M}\sum_{i=1}^{M}{\zeta_i}$, where $\zeta_i$  is bounded: ${\Vert \mathbf{\Sigma}^{-1/2}(\zeta_i - \mathbf{\mu}) \Vert}_2^2 \le R^2, \forall i$.
Then $\hat{\mathbf{\mu}}$ satisfies the following constraint with probability at least $1-\delta_1$:
\[
(\mathbf{\mu} - \hat{\mathbf{\mu}})^T \mathbf{\Sigma}^{-1} (\mathbf{\mu} - \hat{\mathbf{\mu}}) \le \beta(\delta),
\]
where $\beta(\delta_1)=\frac{R^2}{M}\bigg( 2 + \sqrt{2\log{\frac{1}{\delta_1}}} \bigg)^2$.
\end{customlemma}
\begin{proof}
Apply a standadization to the $\zeta_i$, which results in a new random variable $\gamma_i:=\mathbf{\Sigma}^{-1/2}(\zeta_i - \mathbf{\mu})$. It is clear that $\gamma_i$ satisfies Lemma~\ref{th:inequality}. Let $\beta(\delta_1)=\frac{R^2}{M}\bigg( 2 + \sqrt{2\log{\frac{1}{\delta_1}}} \bigg)^2$, then we have:
\begin{equation*}
\begin{aligned}
\mathbb{P}\bigg( (\hat{\mathbf{\mu}}-\mathbf{\mu})^T\mathbf{\Sigma}^{-1}(\hat{\mathbf{\mu}}-\mathbf{\mu}) \le \beta(\delta_1)  \bigg) & = \mathbb{P}\Bigg( \left\Vert \mathbf{\Sigma}^{-1/2}\bigg( \hat{\mathbf{\mu}} - \mathbf{\mu} \bigg)  \right\Vert_2^2 \le \beta(\delta_1) \Bigg) \\
										& = \mathbb{P}\Bigg( \left\Vert \mathbf{\Sigma}^{-1/2}\bigg( \frac{1}{M}\sum_{i=1}^{M}{\zeta_i} - \mathbf{\mu} \bigg)  \right\Vert_2^2 \le \beta(\delta_1)  \Bigg) \\
										& = \mathbb{P}\Bigg(\left\Vert \frac{1}{M}  \sum_{i=1}^{M}{ \mathbf{\Sigma}^{-1/2} \bigg( \zeta_i - \mathbf{\mu}  \bigg) }  \right\Vert_2^2 \le \beta(\delta_1)   \Bigg) \\
										& = \mathbb{P}\Bigg(\frac{1}{M} \left\Vert \sum_{i=1}^{M}{\gamma_i} \right\Vert_2^2 \le \beta(\delta_1) \Bigg) \ge 1 - \delta_1
\end{aligned}
\end{equation*} 
\end{proof}

\section{Proof of Theorem 2}
\begin{customtheorem}{2}
With probability at least $1-\delta$, where $\delta=\delta_1+\delta_2$,  the uncertainty set $\Pi$ contains the ground-truth distribution $\mathcal{P}$.
\end{customtheorem}
\vspace{-0.1in}
\begin{proof}
We define two events $A_1$ and $A_2$ as follow:
\begin{equation*}
\begin{aligned}
& A_1: \text{(C1) holds given $\hat{\mathbf{\Sigma}}$ is close  to $\mathbf{\Sigma}$} \\
& A_2: \text{(C2) holds given $\hat{\mathbf{\mu}}$ is close  to $\mathbf{\mu}$} \\
\end{aligned}
\end{equation*}

Then we have:
\begin{equation*}
\begin{aligned}
\mathbb{P}\big( A_1 \cap A_2 \big)  & = \mathbb{P}\bigg( (A^c_1 \cup A^c_2)^c \bigg) \\
				    & 1 - \mathbb{P}\big( A^c_1 \cup A^c_2 \big) \\									
				    & \quad \text{(by union bound)} \\										   
				    & \ge 1 - \big[\mathbb{P}\big( A^c_1 \big) + \mathbb{P}\big( A^c_1 \big)\big] \\				
				    & \ge 1 - \big[ \delta_1 + \delta_2 \big] \\								
				    & =  1 - \delta,
\end{aligned}
\end{equation*}
where $A_1 \cap A_2$ is the event that $\mathcal{P} \in \Pi$. In other words, $\mathbb{P}\big( \mathcal{P} \in \Pi \big) \ge 1 - \delta$, which completes the proof.
\end{proof}

\section{Detailed dependency of $\mathbf{B}(\mathbf{\mu}), \mathbf{P}(\mathbf{\mu}, \mathbf{\Sigma}), \mathbf{M}(\mathbf{\mu}, \mathbf{\Sigma})$ on their arguments}
In the following we expand the definition of $\mathbf{B}(\mathbf{\mu}), \mathbf{P}(\mathbf{\mu}, \mathbf{\Sigma}), \mathbf{M}(\mathbf{\mu}, \mathbf{\Sigma})$, which makes their dependency on $\mathbf{\mu}$ and $ \mathbf{\Sigma}$ clear:
\begin{equation*}
\begin{aligned}
\mathbf{B}(\mathbf{\mu})  :&= diag\big( \mathbb{E}_{\mathbf{\pi} \sim \mathcal{P}}[\bar{\mathbf{\pi}}] \big) \\ 
		     & = diag\big( 1 - \mathbf{\mu} \big) \\ 
\mathbf{P}(\mathbf{\mu}, \mathbf{\Sigma})  :&= \mathbf{A} \odot \mathbb{E}_{\mathbf{\pi} \sim \mathcal{P}}[\bar{\mathbf{\pi}} \bar{\mathbf{\pi}}^T] \\
             &= \mathbf{A} \odot \bigg( \mathbf{J}(N, N) - \mathbf{J}(N, 1)\times \mathbf{\mu}^T - \mathbf{\mu} \times \mathbf{J}(1, N) + \mathbf{\Sigma} + \mathbf{\mu} \times \mathbf{\mu}^T \bigg) \\
\mathbf{M}(\mathbf{\mu}, \mathbf{\Sigma})  :&= \mathbf{A} \odot \mathbb{E}_{\mathbf{\pi} \sim \mathcal{P}}[ \mathbf{\pi} \bar{\mathbf{\pi}}^T]  \\
             &= \mathbf{A} \odot \bigg( \mathbf{\mu} \times \mathbf{J}(1, N) - \mathbf{\Sigma} - \mathbf{\mu} \times \mathbf{\mu}^T \bigg)
\end{aligned}
\end{equation*}

\section{Detailed forms of the matrices $\mathbf{Q}(\mathbf{\mu}, \mathbf{\Sigma})$ and $\mathbf{b}(\mathbf{\mu})$}
The matrices $\mathbf{Q}(\mathbf{\mu}, \mathbf{\Sigma})$ and $\mathbf{b}(\mathbf{\mu})$ defined in the paper have the following forms:
\begin{equation*}
\begin{aligned}
\mathbf{Q}(\mathbf{\mu}, \mathbf{\Sigma}) :&= \frac{\alpha_3\bigg(\mathbf{M}(\mathbf{\mu}, \mathbf{\Sigma}) + \mathbf{M}(\mathbf{\mu}, \mathbf{\Sigma})^T   \bigg)}{2} - \frac{\alpha_2\bigg( \mathbf{P}(\mathbf{\mu}, \mathbf{\Sigma}) + \mathbf{P}(\mathbf{\mu}, \mathbf{\Sigma})^T  \bigg)}{2} \\
									      & = \alpha_3 \mathbf{A} \odot \Bigg[ \frac{\mathbf{\mu}\mathbf{1}^T+\mathbf{1}\mathbf{\mu}^T}{2} - {\mathbf{\Sigma}} - \mathbf{\mu}\mathbf{\mu}^T \Bigg] -\alpha_2 \mathbf{A}\odot \Bigg[ J(N, N) - \mathbf{1}\mathbf{\mu}^T - \mathbf{\mu}\mathbf{1}^T + {\mathbf{\Sigma}} + \mathbf{\mu}\mathbf{\mu}^T \Bigg]            \\
									      & = \mathbf{A}\odot\Bigg[ \bigg(\frac{\alpha_3+2\alpha_2}{2}\bigg)\mathbf{\mu}\times\mathbf{1}^T + \bigg(\frac{\alpha_3+2\alpha_2}{2}\bigg)\mathbf{1}\times \mathbf{\mu}^T - (\alpha_2+\alpha_3)\mathbf{\Sigma} - (\alpha_2+\alpha_3)\mathbf{\mu}\times\mathbf{\mu}^T - \alpha_2J(N, N) \Bigg] \\
									      & = \mathbf{A} \odot \Bigg[ \bigg(\frac{\alpha_3+2\alpha_2}{2}\bigg)(\mathbf{\mu}-\mathbf{1})\times \mathbf{1}^T + \bigg(\frac{\alpha_3+2\alpha_2}{2}\bigg)\mathbf{1}\times (\mathbf{\mu}-\mathbf{1})^T + (\alpha_2+\alpha_3)J(N, N)\\
									      & - (\alpha_2 + \alpha_3)\mathbf{\Sigma} - (\alpha_2+\alpha_3)\mathbf{\mu}\times\mathbf{\mu}^T \Bigg] \\
\mathbf{b}(\mathbf{\mu})                         :& = (\alpha_1 / 2) \mathbf{B}(\mathbf{\mu}) \mathbf{1} \\
										          & =  (\alpha_1 / 2)(\mathbf{1} - \mathbf{\mu})
\end{aligned}
\end{equation*}

\section{Detailed reformulation of Eq.~(7)}
In the paper in order to apply the \textit{S-Lemma} to convert the two infinite dimensional constraints, Eq.~(6b) and Eq.~(6c), to a finite dimensional constraint, we need two functions in quadratic forms. Notice that Eq.~(6b) is already a quadratic function in $\mathbf{\mu}_\mathcal{F}$. So what remains is to convert Eq.~(6c) to a quadratic function in $\mathbf{\mu}_\mathcal{F}$. We first convert the following to a quadratic function in $\mathbf{\mu}_\mathcal{F}$:
\begin{equation*}
\mathbf{x}^T \mathbf{Q}(\mathbf{\mu}_{\mathcal{F}}, \hat{\mathbf{\Sigma}}) \mathbf{x} +  2 \mathbf{x}^T \mathbf{b}(\mathbf{\mu}_{\mathcal{F}})
\end{equation*}
From last section we know:
\begin{equation}\label{eq:Q}
\begin{aligned}
 \mathbf{Q}(\mathbf{\mu}_{\mathcal{F}}, \hat{\mathbf{\Sigma}}) &=  \bigg(\frac{\alpha_3+2\alpha_2}{2}\bigg)\underbrace{\Bigg[ \mathbf{A} \odot \bigg( (\mathbf{\mu}_\mathcal{F}-\mathbf{1})\times \mathbf{1}^T \bigg) \Bigg]}_{\circled{1}} +  \bigg(\frac{\alpha_3+2\alpha_2}{2}\bigg) \underbrace{\Bigg[ \mathbf{A} \odot \bigg( \mathbf{1}\times (\mathbf{\mu}_\mathcal{F}-\mathbf{1})^T\bigg) \Bigg]}_{\circled{2}} + \\
& \underbrace{(\alpha_2+\alpha_3) \mathbf{A} - (\alpha_2+\alpha_3)\big(\mathbf{A} \odot \hat{\mathbf{\Sigma}} \big) - (\alpha_2 + \alpha_3)\mathbf{A} \odot \big(\mathbf{\mu}_\mathcal{F}\times\mathbf{\mu}_\mathcal{F}^T \big)}_{\circled{3}}
\end{aligned}
\end{equation}
The three terms \circled{1}, \circled{2} and \circled{3}, together with $\mathbf{x}$, form three quadratic functions in $\mathbf{x}$. In what follows, we convert them to quadratic functions in $\mathbf{\mu}_\mathcal{F}$. Note that the operator $diag(\mathbf{x})$ returns a diagonal matrix with diagonal elements equal to $\mathbf{x}$:
\begin{equation*}
\begin{aligned}
\mathbf{x}^T \underbrace{\Bigg[ \mathbf{A} \odot \bigg( \mathbf{1} \times (\mathbf{\mu}_\mathcal{F}-\mathbf{1})^T \bigg) \Bigg]}_{\circled{2}}\mathbf{x} & \stackrel{(\ast)}{=} Tr\Bigg[ diag(\mathbf{x}) \cdot \mathbf{A} \cdot diag(\mathbf{x}) \cdot (\mathbf{\mu}_\mathcal{F} - \mathbf{1})\times\mathbf{1}^T \Bigg] = \\
&  \color{blue}\text{(the trace operator is invariant under cyclic permutations)} \\
& = Tr\Bigg[ \mathbf{A} \cdot diag(\mathbf{x}) \cdot (\mathbf{\mu}_\mathcal{F} - \mathbf{1})\times\mathbf{1}^T \cdot diag(\mathbf{x}) \Bigg]  \\
& = Tr\Bigg[ \mathbf{A} \cdot  diag(\mathbf{x}) \cdot (\mathbf{\mu}_\mathcal{F} - \mathbf{1})\cdot \mathbf{x}^T \Bigg] \\
& \quad \color{blue}\text{$\Big(Tr\big[\mathbf{A}\cdot diag(\mathbf{x}) \cdot \mathbf{1} \cdot \mathbf{x}^T\big]= Tr\big[ \mathbf{A} \mathbf{x} \mathbf{x}^T \big] = \mathbf{x}^T\mathbf{A} \mathbf{x}\Big)$} \\
& = Tr\Bigg[ \mathbf{A} \cdot diag(\mathbf{x}) \cdot \mathbf{\mu}_\mathcal{F} \cdot \mathbf{x}^T \Bigg] - \mathbf{x}^T \mathbf{A} \mathbf{x} \\
& \quad \color{blue} \text{$\Big(diag(\mathbf{x}) \mathbf{\mu}_\mathcal{F} = diag(\mathbf{\mu}_\mathcal{F})\mathbf{x} \Big)$} \\
& = Tr\Bigg[ \mathbf{A} \cdot diag(\mathbf{\mu}_\mathcal{F})\cdot \mathbf{x}\mathbf{x}^T \Bigg] - \mathbf{x}^T \mathbf{A} \mathbf{x} \\
& = \mathbf{x}^T \Bigg[\mathbf{A} \cdot diag(\mathbf{x})  \Bigg]\mathbf{\mu}_\mathcal{F} - \mathbf{x}^T \mathbf{A} \mathbf{x} 
\end{aligned}
\end{equation*}
where $(\ast)$ comes from the fact that $\mathbf{x}^T[\mathbf{A} \odot \mathbf{B}]\mathbf{x} = Tr[diag(\mathbf{x}) \cdot \mathbf{A} \cdot diag(\mathbf{x}) \cdot \mathbf{B}^T]$. Similarly we have:
\begin{equation*}
\begin{aligned}
\mathbf{x}^T \underbrace{\Bigg[ \mathbf{A} \odot \bigg( (\mathbf{\mu}_\mathcal{F}-\mathbf{1})\times \mathbf{1}^T \bigg) \Bigg]}_{\circled{1}}\mathbf{x} & = Tr\Bigg[ diag(\mathbf{x}) \cdot \mathbf{A} \cdot diag(\mathbf{x}) \times \mathbf{1} \times (\mathbf{\mu}_\mathcal{F} - \mathbf{1})^T  \Bigg] \\
& = Tr\Bigg[diag(\mathbf{x}) \cdot \mathbf{A} \cdot \mathbf{x} \cdot (\mathbf{\mu}_\mathcal{F} - \mathbf{1})^T \Bigg] \\
& = Tr\Bigg[ diag(\mathbf{x}) \cdot \mathbf{A} \cdot \mathbf{x} \cdot \mathbf{\mu}_\mathcal{F} \Bigg] - \mathbf{x}^T \mathbf{A} \mathbf{x} \\
& = \mathbf{x}^T \Bigg[\mathbf{A} \cdot diag(\mathbf{x})  \Bigg]\mathbf{\mu}_\mathcal{F} - \mathbf{x}^T \mathbf{A} \mathbf{x} \\
\end{aligned}
\end{equation*}
and:
\begin{equation*}
\begin{aligned}
& \mathbf{x}^T \underbrace{\Bigg[(\alpha_2+\alpha_3) \mathbf{A} - (\alpha_2+\alpha_3)\big(\mathbf{A} \odot \hat{\mathbf{\Sigma}} \big)  - (\alpha_2 + \alpha_3)\mathbf{A} \odot \big(\mathbf{\mu}_\mathcal{F}\mathbf{\mu}_\mathcal{F}^T \big)\Bigg]}_{\circled{3}} \mathbf{x}  = \\
&  (\alpha_2+\alpha_3)\mathbf{x}^T \mathbf{A} \mathbf{x} - (\alpha_2+\alpha_3) \mathbf{x}^T \big(\mathbf{A} \odot \hat{\mathbf{\Sigma}} \big)\mathbf{x} - \underbrace{(\alpha_2 + \alpha_3) \mathbf{\mu}_\mathcal{F}^T\bigg(\mathbf{A}\odot \big(\mathbf{x}\mathbf{x}^T \big) \bigg) \mathbf{\mu}_\mathcal{F}}_{(\diamond)},
\end{aligned}
\end{equation*}
where $(\diamond)$ comes from the following:
\begin{equation*}
\begin{aligned}
& -(\alpha_2+\alpha_3) \mathbf{x}^T \bigg[ \mathbf{A} \odot \big( \mathbf{\mu}_\mathcal{F} \mathbf{\mu}_\mathcal{F}^T \big) \bigg] \mathbf{x} \\
& = -(\alpha_2+\alpha_3) Tr\Bigg[ diag(\mathbf{x}) \cdot \mathbf{A} \cdot diag(\mathbf{x}) \cdot \bigg( \mathbf{\mu}_\mathcal{F} \mathbf{\mu}_\mathcal{F}^T   \bigg) \Bigg] \\
& =  -(\alpha_2+\alpha_3) Tr\Bigg[ diag(\mathbf{x}) \cdot \mathbf{A} \cdot diag(\mathbf{\mu}_\mathcal{F}) \cdot \mathbf{x} \cdot \mathbf{\mu}_\mathcal{F}^T \Bigg] \\
& = -(\alpha_2+\alpha_3) Tr\Bigg[  \mathbf{A} \cdot diag(\mathbf{\mu}_\mathcal{F}) \cdot \mathbf{x} \cdot \mathbf{\mu}_\mathcal{F}^T \cdot diag(\mathbf{x})\Bigg] \\
& =  -(\alpha_2+\alpha_3) Tr\Bigg[ \mathbf{A} \cdot diag(\mathbf{\mu}_\mathcal{F}) \cdot \mathbf{x} \mathbf{x}^T \cdot diag(\mathbf{\mu}_\mathcal{F}) \Bigg] \\
& = -(\alpha_2+\alpha_3) Tr\Bigg[  diag(\mathbf{\mu}_\mathcal{F}) \cdot \mathbf{x} \mathbf{x}^T \cdot diag(\mathbf{\mu}_\mathcal{F}) \cdot  \mathbf{A} \Bigg] \\
& =  - (\alpha_2 + \alpha_3) \mathbf{\mu}_\mathcal{F}^T\bigg(\mathbf{A}\odot \big(\mathbf{x}\mathbf{x}^T \big) \bigg) \mathbf{\mu}_\mathcal{F}
\end{aligned}
\end{equation*}
Putting the above derivation together we obtain:
\begin{equation*}
\begin{aligned}
& \mathbf{x}^T \mathbf{Q}(\mathbf{\mu}_{\mathcal{F}}, \hat{\mathbf{\Sigma}}) \mathbf{x} +  2 \mathbf{x}^T \mathbf{b}(\mathbf{\mu}_{\mathcal{F}}) = \\
\mathbf{\mu}_{\mathcal{F}}^T & \Bigg[ - (\alpha_2 + \alpha_3) \bigg(\mathbf{A}\odot \big(\mathbf{x}\mathbf{x}^T \big) \bigg)  \Bigg] \mathbf{\mu}_{\mathcal{F}} + \mathbf{\mu}_\mathcal{F} ^T \Bigg[ (\alpha_3+2\alpha_2)   diag(\mathbf{x})\cdot \mathbf{A} \cdot \mathbf{x}  - \alpha_1\mathbf{x} \Bigg] - \\
&\Bigg[ \alpha_1 \mathbf{1}^T\mathbf{x} - \mathbf{x}^T\bigg(  (\alpha_2+\alpha_3)\big( \mathbf{A}\odot \hat{\mathbf{\Sigma}} \big) + \alpha_2 \mathbf{A}   \bigg) \mathbf{x}  \Bigg]
\end{aligned}
\end{equation*}
So the function $f(\mathbf{\mu}_\mathcal{F})$ becomes:
\begin{equation*}
\begin{aligned}
& \mathbf{x}^T \mathbf{Q}(\mathbf{\mu}_{\mathcal{F}}, \hat{\mathbf{\Sigma}}) \mathbf{x} +  2 \mathbf{x}^T \mathbf{b}(\mathbf{\mu}_{\mathcal{F}}) - t - \mathbf{\mu}_{\mathcal{F}}^T \mathbf{K} \mathbf{\mu}_{\mathcal{F}}  = \\
& \mathbf{\mu}_{\mathcal{F}}^T \Bigg[ - (\alpha_2 + \alpha_3) \bigg(\mathbf{A}\odot \big(\mathbf{x}\mathbf{x}^T \big) \bigg) - \mathbf{K}  \Bigg] \mathbf{\mu}_{\mathcal{F}} + \mathbf{\mu}_\mathcal{F} ^T \Bigg[ (\alpha_3 + 2\alpha_2)diag(\mathbf{x})\cdot \mathbf{A} \cdot \mathbf{x}  - \alpha_1 \mathbf{x}   \Bigg] + \\
& \qquad \Bigg[ \alpha_1 \mathbf{1}^T\mathbf{x} - \mathbf{x}^T\bigg(  (\alpha_2+\alpha_3)\big( \mathbf{A}\odot \hat{\mathbf{\Sigma}} \big) + \alpha_2 \mathbf{A}   \bigg) \mathbf{x} - t \Bigg],
\end{aligned}
\end{equation*}
which is a quadratic function in $\mathbf{\mu}_\mathcal{F}$. Define $\mathbf{R}, \mathbf{r}$ and $z$ as the following:
\begin{equation*}
\begin{aligned}
& \mathbf{R}  =  - (\alpha_2 + \alpha_3) \bigg(\mathbf{A}\odot \big(\mathbf{x}\mathbf{x}^T \big) \bigg) - \mathbf{K} \\
& \mathbf{r}  = (\alpha_3+2\alpha_2)diag(\mathbf{x})\cdot \mathbf{A} \cdot \mathbf{x}  - \alpha_1 \mathbf{x}   \\
& z           = \alpha_1 \mathbf{1}^T\mathbf{x} - \mathbf{x}^T\bigg(  (\alpha_2+\alpha_3)\big( \mathbf{A}\odot \hat{\mathbf{\Sigma}} \big) + \alpha_2 \mathbf{A}   \bigg) \mathbf{x} - t,
\end{aligned}
\end{equation*}
which results in a compact form of $f(\mathbf{\mu}_\mathcal{F})$:
\begin{equation*}
f(\mathbf{\mu}_\mathcal{F}) = \mathbf{\mu}_{\mathcal{F}}^T \mathbf{R} \mathbf{\mu}_{\mathcal{F}} + \mathbf{\mu}_{\mathcal{F}}^T \mathbf{r} + z
\end{equation*}

\section{Proof of Eq.(10) in the paper}
The relation $(r1)$ is direct. To see why $(r2)$ holds, note that the $i$-th element of $diag(\mathbf{x})\cdot \mathbf{A} \cdot \mathbf{x}$ is:
\begin{equation*}
\Big[ diag(\mathbf{x})\cdot \mathbf{A} \cdot \mathbf{x} \Big]_i = \mathbf{x}_i \sum_{j=1}^{N}{\mathbf{A}_{ij}\mathbf{x}_j},
\end{equation*}
which is equal to the $i$-th element of $diag(\mathbf{A}\mathbf{X})$:
\begin{equation*}
\Big[ diag(\mathbf{A}\mathbf{X})\Big]_i = \Big[diag(\mathbf{A})\mathbf{x}\mathbf{x}^T \Big]_i = \mathbf{x}_i \sum_{j=1}^{N}{\mathbf{A}_{ij}\mathbf{x}_j}.
\end{equation*}
The relation $(r3)$ holds because:
\begin{equation*}
\begin{aligned}
\mathbf{x}^T \bigg(   (\alpha_2+\alpha_3)(\mathbf{A}\odot \hat{\mathbf{\Sigma}}) + \alpha_2\mathbf{A} \bigg) \mathbf{x} & = (\alpha_2+\alpha_3) \mathbf{x}^T (\mathbf{A}\odot \hat{\mathbf{\Sigma}}) \mathbf{x} + \alpha_2 \mathbf{x}^T \mathbf{A}\mathbf{x}  \\
& =  (\alpha_2+\alpha_3) Tr\Big[(\mathbf{A}\odot \hat{\mathbf{\Sigma}}) \mathbf{x} \mathbf{x}^T  \Big] + \alpha_2 Tr\Big[\mathbf{A} \mathbf{x}\mathbf{x}^T  \Big] \\
& =   (\alpha_2+\alpha_3) Tr\Big[(\mathbf{A}\odot \hat{\mathbf{\Sigma}}) \mathbf{X} \Big] + \alpha_2 Tr\Big[\mathbf{A} \mathbf{X} \Big]
\end{aligned}
\end{equation*}

\break
\section{Statistics of the networks used in experiments}
\begin{table*}[h]
\centering
\footnotesize
\begin{tabular}{ccccc}
\toprule
		& $r$          & density & \#edges & clustering coeff. \\
\midrule
\textit{BA-1}            & 2.7167       & 0.0461  & 375     & 0.1340 \\
\textit{BA-2}            & 2.2789       & 0.0610  & 496     & 0.1504 \\
\textit{BA-3}            & 2.0374       & 0.0757  & 615     & 0.1646 \\
\textit{SW-1}            &              & 0.0787  & 640     & 0.3664 \\
\textit{SW-2}            &              & 0.1102  & 896     & 0.3875 \\
\textit{SW-3}            &              & 0.1575  & 1280    & 0.4059 \\
Facebook        &              & 0.0106  & 1325    & 0.3930 \\
\bottomrule
\end{tabular}
\caption{Statistics of networks used in our experiments. $r$ is the exponent of the power-law degree distribution.}
\end{table*}

\end{document}